\documentclass{article} 
\usepackage{colm2024_conference}

\usepackage{amsmath,amsfonts,bm}

\def\eqref#1{equation~\ref{#1}}

\def\1{\bm{1}}

\DeclareMathAlphabet{\mathsfit}{\encodingdefault}{\sfdefault}{m}{sl}
\SetMathAlphabet{\mathsfit}{bold}{\encodingdefault}{\sfdefault}{bx}{n}

\usepackage[T1]{fontenc}
\usepackage{hyperref}
\usepackage{url}
\usepackage{booktabs} 
\usepackage{multirow}
\usepackage{enumitem}
\usepackage{amsmath} 
\usepackage{natbib}
\hypersetup{colorlinks=true, allcolors=blue}
\usepackage{caption}
\usepackage{adjustbox}
\usepackage{pifont}
\usepackage{listings}
\usepackage{multirow}
\usepackage[table]{xcolor}
\usepackage{arydshln}
\usepackage{xcolor}
\usepackage{titletoc}
\usepackage{subcaption}
\usepackage{wrapfig}
\usepackage{amssymb}
\usepackage[most]{tcolorbox}
\usepackage{xcolor}
\usepackage{tcolorbox}
\usepackage{amssymb}
\usepackage{float}
\usepackage{algorithm}
\usepackage{algorithmic}

\floatname{algorithm}{Algorithm}

\usepackage[most]{tcolorbox}
\usepackage{xcolor}
\usepackage{listings}

\usepackage[table]{xcolor}
\definecolor{bestbg}{RGB}{235,224,252}
\definecolor{bestbg}{RGB}{235,224,252}
\definecolor{secondbg}{RGB}{224,237,253}

\newcommand{\best}[1]{\cellcolor{bestbg}\textbf{#1}}
\newcommand{\second}[1]{\cellcolor{secondbg}\underline{#1}}

\title{ToolOPD: Credit-Aware On-Policy Distillation for Tool-Use Agents}

\author{
Chu Zhao$^{1}$,
Lei Tang$^{2}$,
Yanming Wang$^{2}$,
Jianzhe Zhao$^{1}$,
Guibing Guo$^{1}$\thanks{Corresponding authors.},
Zhengzong Chen$^{2*}$,
Yuanyuan Zhao$^{2}$,
Fei Huang$^{2}$\\
$^{1}$Northeastern University\\
$^{2}$Honor Device Co., Ltd\\
}

\begin{document}
\maketitle
\vspace{-30pt}
\begin{center}
    \ghlink
\end{center}

\begin{abstract}
On-policy distillation (OPD) enhances the capabilities of small language agents by providing dense, token-level supervision from a teacher model on student-generated trajectories. However, standard OPD implicitly assumes that credit is uniformly distributed across long-horizon trajectories, assigning the same strength of teacher supervision to different actions. Through credit-distribution analysis and outcome-reversal counterfactual experiments, we find that trajectory credit is concentrated on a small number of critical actions. Moreover, even when the trajectory is held fixed, reversing only the final outcome substantially changes the credit assigned to multiple interaction turns. These findings show that the uniform-credit assumption fails to identify critical decisions and ultimately degrades the model's reasoning accuracy. To address these challenges, we propose \textbf{ToolOPD}, an outcome-aware on-policy distillation framework  designed for long-horizon tool use.  ToolOPD first summarizes each completed trajectory and
its outcome into a reusable hindsight skill. It then measures how this skill changes the teacher's probability of each student action token and converts the change into outcome-aware credit. The resulting credit preserves supervision that supports successful decisions or discourages failed ones, while suppressing irrelevant and conflicting signals. ToolOPD uses these credits to reweight the action-level distillation loss and jointly optimizes it with reinforcement learning. Extensive experiments across multiple long-horizon tool interaction benchmarks demonstrate that ToolOPD significantly outperforms existing reinforcement learning and self-distillation baselines.
\end{abstract}

\section{Introduction}

\begin{figure*}[ht]
    \centering
    \includegraphics[width=0.24\linewidth]{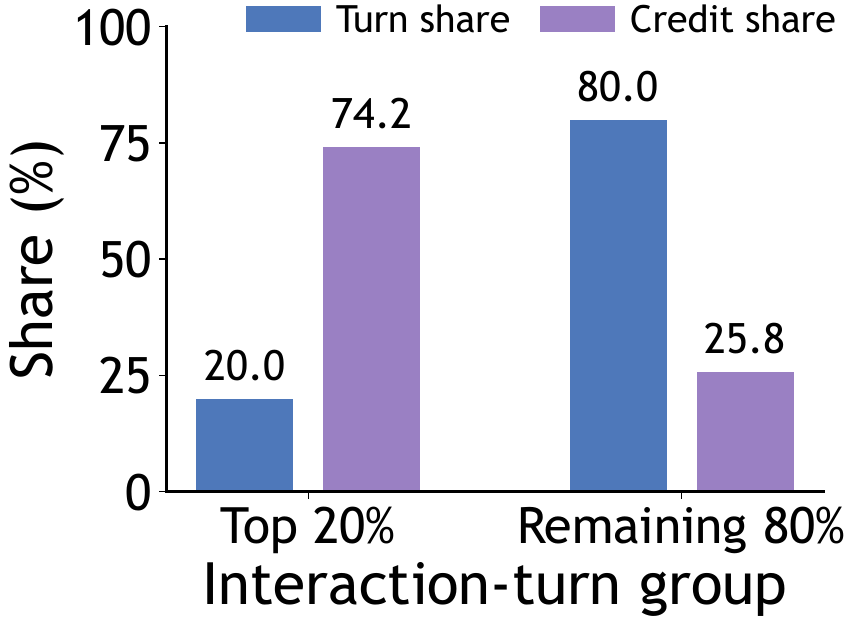}
    \hfill
    \includegraphics[width=0.24\linewidth]{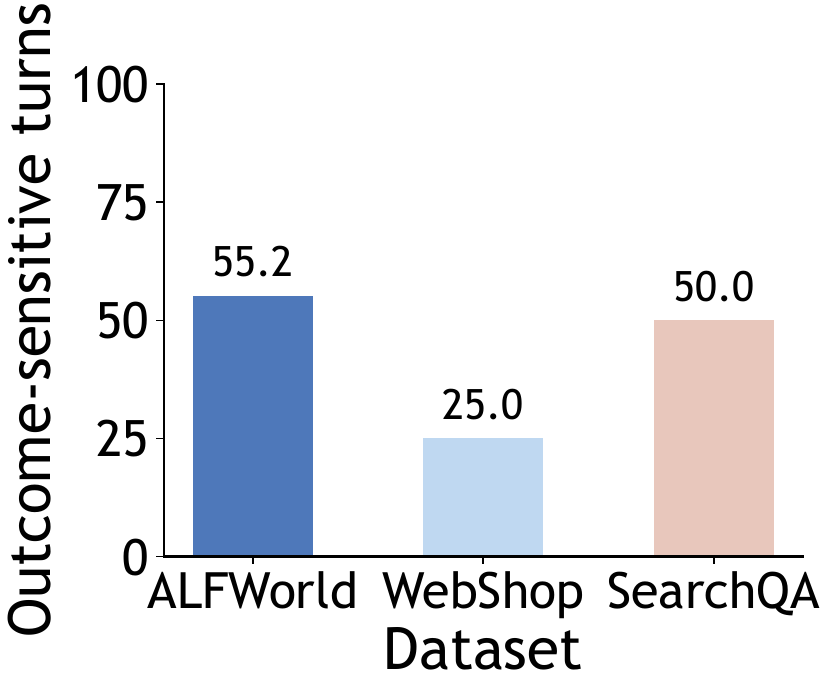}
    \hfill
    \includegraphics[width=0.24\linewidth]{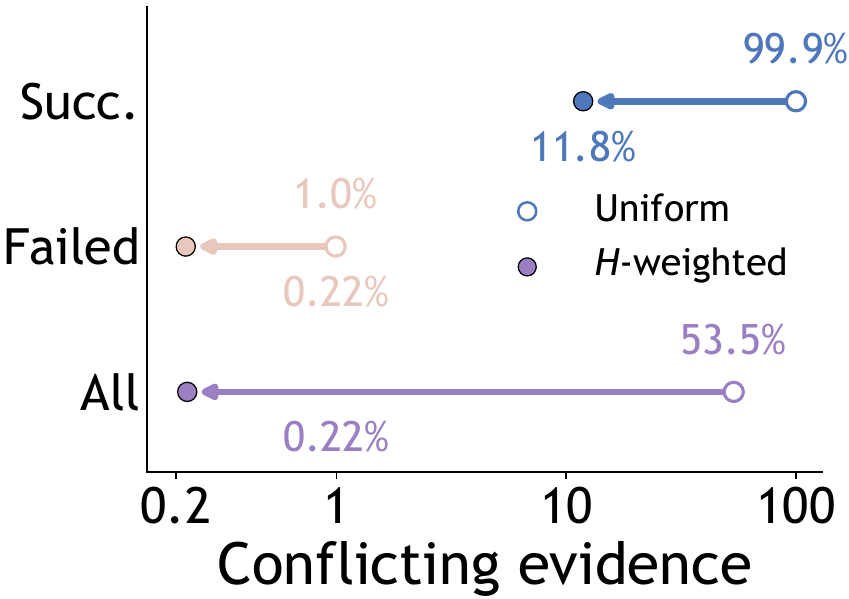}
    \hfill
    \includegraphics[width=0.24\linewidth]{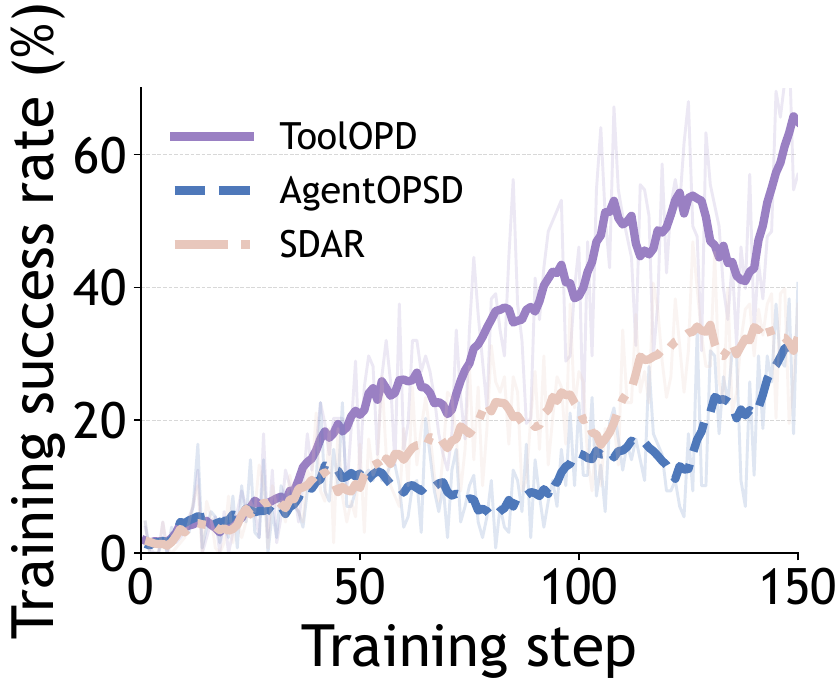}

    \caption{Empirical Study Results.}
    \label{fig:motivation}
\end{figure*}

Large language models (LLMs)~\citep{naveed2025comprehensive,minaee2024large,chang2024survey,zhao2026survey,thirunavukarasu2023large}
have demonstrated strong capabilities in complex interactive tasks by invoking
external tools~\citep{qu2025tool,xu2025llm,qin2024tool} and adapting their
decisions to environmental feedback. However, long-horizon tool-use
tasks~\citep{liu2026planbench,li2026phgpo} often require tens or hundreds of
interaction turns while providing only a sparse success-or-failure reward at the
end of the trajectory. This makes it difficult to determine which intermediate
actions contribute to the final outcome. On-policy distillation
(OPD)~\citep{agarwal2024policy,li2026rethinking,song2026survey,yang2026learning}
addresses sparse supervision by matching a compact student policy to a teacher
distribution on trajectories generated by the student itself. Standard OPD implicitly assumes that credit is uniformly distributed across long-horizon trajectories, such that different actions receive equal credit and the same strength of teacher supervision. This uniform credit assignment fails to distinguish critical tool decisions from routine transitional actions, thereby weakening effective supervision for critical decisions.

To systematically characterize this issue, we conduct a series of empirical studies.
As shown in Fig.~\ref{fig:motivation}(a), on the ALFWorld diagnostic
trajectories, the top 20\% of interaction turns account for 74.2\% of the total
credit, demonstrating that trajectory credit is highly concentrated rather than
uniformly distributed. We further construct an outcome-reversal counterfactual
experiment by fixing the complete task and state--action sequence while reversing
only the outcome label provided to the teacher. Fig.~\ref{fig:motivation}(b)
shows that this intervention substantially changes the credit assigned to
55.2\%, 25.0\%, and 50.0\% of interaction turns in ALFWorld, WebShop, and
SearchQA, respectively. Thus, the credit of an action depends not only on its
local context but also on its relationship with the final outcome.
Fig.~\ref{fig:motivation}(c) further shows that outcome-conflicting signals
account for 53.5\% of the total teacher evidence under uniform distillation.
Together, these findings establish that effective long-horizon distillation must
identify critical actions and determine whether the corresponding teacher
signals are consistent with the trajectory outcome. Detailed settings are
provided in Appendix~\ref{ap:empirical_study_settings}.
Recent studies have begun to explore fine-grained credit assignment for long-horizon tasks. ATOD~\citep{tan2026atod} adaptively adjusts the distillation strength of different actions according to the discrepancy between the teacher and student policies, while TurnSight~\citep{qu2026turnsight} constructs multi-view hindsight information from actual tool executions and selects reliable turn-level supervision through cross-view consistency. 

Unlike existing
methods that determine distillation weights from local teacher--student
divergence, training progress, or interaction difficulty,
in this work, we propose \textbf{ToolOPD}, an outcome-aware
on-policy distillation framework for long-horizon tool use. ToolOPD uses the
terminal outcome to determine both the direction and strength of action credit.
For each student-generated trajectory, ToolOPD conditions the teacher on the
complete interaction process and its outcome to induce a concise hindsight
skill that summarizes the effective workflow or failure pattern. It then
compares the teacher's action probabilities with and without the hindsight skill
to estimate the outcome-relevant credit of each action token. For successful
trajectories, ToolOPD retains skill-induced probability increases that reinforce
effective actions; for failed trajectories, it retains probability decreases
that suppress erroneous actions. The resulting relevance is further calibrated
by teacher confidence and experience-based reliability to prevent uncertain or
historically inconsistent teacher signals from dominating the update. ToolOPD
uses the calibrated credit to dynamically reweight the distillation loss over
\texttt{<action>} tokens and jointly optimizes it with Group Relative Policy
Optimization (GRPO)~\citep{guo2025deepseek}. This design concentrates supervision on outcome-relevant
decisions while filtering redundant and conflicting signals. 

Our main contributions are summarized as follows:
\begin{itemize}[left=0pt]
    \item \textbf{Empirical findings.}
    We show that credit in long-horizon OPD is concentrated on critical actions and depends on trajectory outcomes, while uniform distillation transfers outcome-conflicting supervision.

    \item \textbf{Method.}
    We propose ToolOPD, which induces outcome-conditioned hindsight skills and converts their effects on teacher predictions into fine-grained, outcome-asymmetric action credit.

    \item \textbf{Experiments.}
    Across multiple tool-use benchmarks and model scales, ToolOPD consistently outperforms representative reinforcement learning and OPD baselines in task performance, long-context robustness, and optimization stability.
\end{itemize}

\section{Method}

\subsection{Problem Formulation}
We consider multi-turn tool use tasks with a discrete set of executable actions. For a given task instance $x_i$, the student policy $\pi_\theta$ interacts with the external environment and generates a trajectory:
\begin{equation}
    \tau_i = \left\{ (s_{i,t}, a_{i,t}, o_{i,t+1}) \right\}_{t=0}^{T_i-1},
\end{equation}
where $s_{i,t}$ denotes the interaction history and the current environmental state at step $t$, $a_{i,t} \in \mathcal{A}(s_{i,t})$ is the action selected by the student policy, and $o_{i,t+1}$ is the observation feedback returned by the environment. Upon the termination of the trajectory, the environment returns a scalar task reward $R_i$ and a binary outcome indicator $y_i = \mathbb{I}(R_i > 0) \in \{0, 1\}$.
In long-horizon tool use scenarios, this setting introduces severe credit assignment challenges: due to the extremely long interaction horizon and highly sparse rewards, the success or failure of a task is typically determined by only a few critical tool decisions (e.g., specific API selections or parameter settings), while the vast majority of the remaining turns consist of routine state queries and transitional actions. How to accurately attribute and assign the terminal global reward to the intermediate specific actions is the core difficulty in optimizing such policies.

We are given a teacher model $\pi_\phi$ with frozen parameters. Our objective is to distill the transferable strategy summarized by the teacher based on the complete trajectory and the final outcome into the student, while preserving the exploration capability of on-policy reinforcement learning. The trajectories during training are generated entirely by the current student policy, i.e., $\tau_i \sim \pi_\theta$. Therefore, our method falls under the category of On-Policy Distillation, rather than simple behavior cloning on offline teacher trajectories.

\subsection{Overview of ToolOPD}
\begin{figure*}[t]
  \centering
  \includegraphics[width=\linewidth]{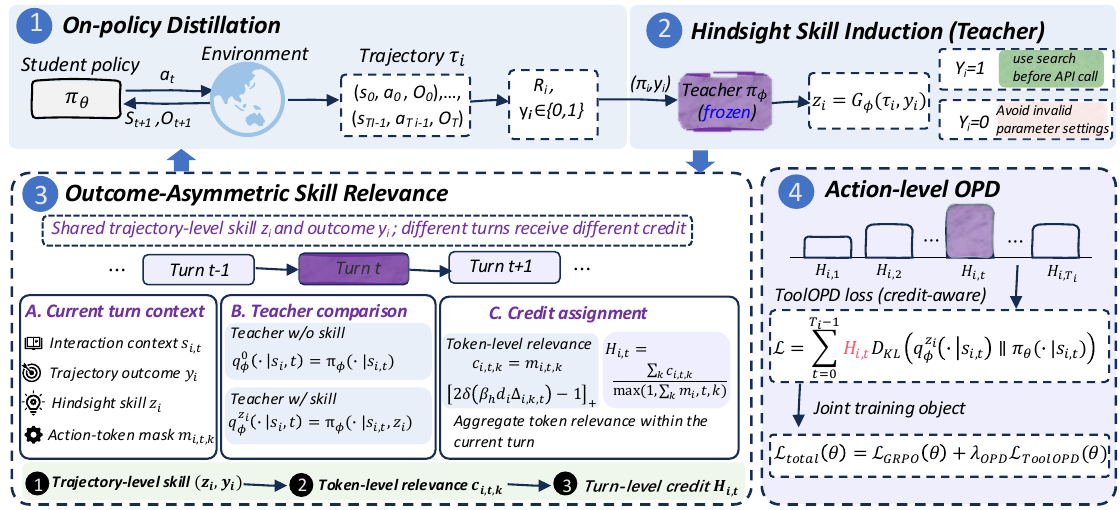}
  \caption{ToolOPD framework overview.}
  \label{fig:toolopd}
\end{figure*}

To address the credit assignment dilemma caused by sparse rewards in long-horizon interactions, ToolOPD organically integrates GRPO with outcome-aware on-policy distillation. Its core idea is to transform sparse trajectory-level credit into dense action-level credit. The specific data flow is as follows: First, the student policy interacts with the environment to generate a set of on-policy trajectories, and the intra-group relative advantages for GRPO are computed. Subsequently, the teacher reads the complete trajectory and the final success/failure outcome to induce a concise, reusable hindsight skill. Next, by comparing the probability distributions of the teacher with and without this skill on the student's actions, the outcome relevance of the skill to the current action is calculated, thereby achieving fine-grained credit assignment. Finally, a weighted teacher-to-student OPD is performed exclusively on the action (\texttt{<action>}) tokens. Throughout the training process, the teacher model parameters remain frozen, and all gradients are used solely to update the student model. The framework of ToolOPD is illustrated in Fig.~\ref{fig:toolopd}.

\subsection{Outcome-Conditioned Hindsight Skill Induction}
Standard distillation typically directly imitates the local action distribution of the teacher, but the teacher lacks global perception of the final success or failure of the current trajectory in this process. To address this issue, ToolOPD first introduces a hindsight skill induction mechanism, allowing the teacher to distill experience using the complete trajectory and the final outcome:
\begin{equation}
    z_i = G_\phi(\tau_i, y_i),
\end{equation}
where $G_\phi$ denotes the skill induction process of the teacher. For successful trajectories ($y_i=1$), the teacher summarizes effective operational procedures or key decisions; for failed trajectories ($y_i=0$), the teacher identifies the root cause of the error and provides a general rule to avoid it. To facilitate context injection, the skill is constrained to a single concise rule, formatted as: \texttt{Rule Name: imperative guidance.}

This design serves a dual purpose: first, it compresses lengthy interaction trajectories into reusable strategy priors, essentially distilling the macroscopic outcome credit at the trajectory level into microscopic strategy signals that can guide specific decisions; second, it prevents the teacher from directly rewriting the complete trajectory, thereby focusing the distillation signal on high-level decision-making patterns. All states within the same trajectory share the corresponding skill $z_i$ for that trajectory. During the inference phase, the student can be deployed independently without the teacher or explicit skill generation, thus incurring no additional inference cost.

\subsection{Outcome-Asymmetric Skill Relevance}
Hindsight skills do not possess equal explanatory power for every decision within a trajectory, which precisely reflects the highly uneven credit distribution in long-horizon interactions. To measure the guiding value of a skill for the current action (i.e., to perform fine-grained credit assignment), we compare the probability distributions of the teacher on the student's actual sampled actions, both with and without the skill.
We define the base teacher distribution and the skill-conditioned teacher distribution as:
\begin{equation}
     q_\phi^0(\cdot \mid s_{i,t}) = \pi_\phi(\cdot \mid s_{i,t}), \\q_\phi^{z_i}(\cdot \mid s_{i,t}) = \pi_\phi(\cdot \mid s_{i,t}, z_i).
\end{equation}
For the $k$-th token $a_{i,t,k}$ in the student's action, we define its skill-induced log-probability shift as:
\begin{equation}
    \Delta_{i,t,k} = \log q_\phi^{z_i}(a_{i,t,k} \mid s_{i,t}) - \log q_\phi^0(a_{i,t,k} \mid s_{i,t}).
\end{equation}
We interpret this change asymmetrically based on the final outcome of the trajectory. Let the direction factor be $d_i = 2y_i - 1 \in \{-1, 1\}$, then the token-level outcome relevance is computed as:
\begin{equation}
    c_{i,t,k} = m_{i,t,k} \left[ 2\sigma\left( \beta_h d_i \Delta_{i,t,k} \right) - 1 \right]_+,
\end{equation}
where $\sigma(\cdot)$ is the sigmoid function, $[\cdot]_+ = \max(\cdot, 0)$, $\beta_h$ controls the mapping sensitivity, and $m_{i,t,k} \in \{0, 1\}$ is the \texttt{<action>} token mask.

This definition embodies outcome asymmetry and achieves dynamic credit assignment: when the trajectory is successful ($y_i=1$), positive relevance is obtained only when the skill increases the probability of the successful action ($\Delta > 0$); when the trajectory fails ($y_i=0$), positive relevance is obtained only when the skill decreases the probability of the failed action ($\Delta < 0$); skill influences inconsistent with the direction of the final outcome are truncated to zero. Averaging the token relevance over the current action yields the decision-level weight:
\begin{equation}
    H_{i,t} = \frac{\sum_k c_{i,t,k}}{\max\left(1, \sum_k m_{i,t,k}\right)}.
\end{equation}
This weight essentially quantifies the credit contribution of the current action to the final outcome. In the subsequent distillation, we directly adopt $w_{i,t} = H_{i,t}$ as the distillation weight for OPD.

\subsection{Action-Level On-Policy Distillation}
The standard OPD objective is to minimize the KL divergence from the teacher distribution to the student distribution over the state distribution generated by the student. Leveraging the dynamically assigned credit weights, ToolOPD can precisely pinpoint critical decisions in long-horizon trajectories and dynamically adjust the distillation intensity for different decisions:
\begin{equation}
   \mathcal{L}_{\mathrm{ToolOPD}}(\theta) = \mathbb{E}_{\tau_i \sim \pi_\theta}  \left[ \sum_{t=0}^{T_i-1} H_{i,t}  D_{\mathrm{KL}} \left( q_\phi^{z_i}(\cdot \mid s_{i,t}) \parallel \pi_\theta(\cdot \mid s_{i,t}) \right) \right].    
\end{equation}
Since the training samples are drawn from the student policy $\pi_\theta$, we employ importance sampling to estimate the aforementioned KL divergence. For the action tokens sampled by the student, we define the log-ratio $\ell_{i,t,k} = \log q_\phi^{z_i}(a_{i,t,k} \mid s_{i,t}) - \log \pi_\theta(a_{i,t,k} \mid s_{i,t})$, and compute the clipped importance weight:
\begin{equation}
    r_{i,t,k} = \exp\left( \operatorname{clip}(\ell_{i,t,k}, -c, c) \right),
\end{equation}
where $c$ is the maximum log-ratio clipping value (set to $c=5$ in this study). The unbiased estimation form of the KL divergence for a single token is $r_{i,t,k} \log r_{i,t,k} - r_{i,t,k} + 1$. Therefore, the practically optimized ToolOPD loss is:
\begin{equation}
     \widehat{\mathcal{L}}_{\mathrm{ToolOPD}} = \frac{1}{N_A} \sum_{i,t,k}  m_{i,t,k} H_{i,t} \left( r_{i,t,k} \log r_{i,t,k} - r_{i,t,k} + 1 \right),  
\end{equation}
where $N_A = \sum_{i,t,k} m_{i,t,k}$ denotes the total number of action tokens in the batch. This design ensures that distillation is applied exclusively to the executable decision tokens within \texttt{<action>}, without interfering with natural language reasoning processes such as \texttt{<think>}, thereby allowing the student to retain its own reasoning expression paradigm.

\subsection{Joint Optimization with GRPO}
To preserve the exploration and policy improvement capabilities driven by environmental rewards, we adopt GRPO as the foundational reinforcement learning objective. For $G$ trajectories sampled for the same task, the group-relative advantage is computed as:
\begin{equation}
    A_i = \frac{R_i - \operatorname{mean}(\{R_j\}_{j=1}^G)}{\operatorname{std}(\{R_j\}_{j=1}^G) + \epsilon}.
\end{equation}
Defining the token-level probability ratio between the current policy and the sampling policy as $\rho_{i,t,k}(\theta) = \frac{\pi_\theta(a_{i,t,k} \mid s_{i,t})}{\pi_{\theta_{\mathrm{old}}}(a_{i,t,k} \mid s_{i,t})}$, the clipped policy loss for GRPO is:
\begin{equation}
     \mathcal{L}_{\mathrm{GRPO}} = -\mathbb{E}_{i,t,k} \left[ \min \left( \rho_{i,t,k} A_i, \operatorname{clip}(\rho_{i,t,k}, 1-\epsilon_l, 1+\epsilon_h) A_i \right) \right]. 
\end{equation}

Ultimately, we combine the reinforcement learning signal from GRPO with the distillation signal from ToolOPD, defining the overall training objective as:
\begin{equation}
    \mathcal{L}_{\mathrm{total}}(\theta) = \mathcal{L}_{\mathrm{GRPO}}(\theta) + \lambda_{\mathrm{OPD}} \widehat{\mathcal{L}}_{\mathrm{ToolOPD}}(\theta),
\end{equation}
where $\lambda_{\mathrm{OPD}}$ is the distillation loss weight. Under this configuration, GRPO is responsible for learning the foundational task policy from environmental rewards and maintaining exploration, while ToolOPD precisely addresses the fine-grained credit assignment problem in long-horizon tool use through outcome-conditioned teacher supervision. The smaller distillation coefficient ensures that the teacher signal serves as an auxiliary regularizer, guiding the direction of policy optimization without overriding the primary gradients of reinforcement learning. The pseudocode of ToolOPD is provided in Appendix~\ref{ap:algorithm}, and the prompt templates used are presented in Appendix~\ref{ap:training_prompts}

\begin{table*}[t]
\centering
\caption{Performance comparison across ALFWorld, Search QA, and WebShop benchmarks.}
\label{tab:horizon_results}
\small
\setlength{\tabcolsep}{3pt}
\renewcommand{\arraystretch}{1.1}

\begin{adjustbox}{max width=\textwidth}
\begin{tabular}{l*{17}{c}}
\toprule
\textbf{Method} &
\multicolumn{7}{c}{\textbf{ALFWorld}} &
\multicolumn{8}{c}{\textbf{Search QA}} &
\multicolumn{2}{c}{\textbf{WebShop}} \\
\cmidrule(lr){2-8}
\cmidrule(lr){9-16}
\cmidrule(lr){17-18}

&
Pick & Look & Clean & Heat & Cool & Pick2 & \textbf{Avg.} &
NQ & Triv & Pop & Hotp & 2Wk & MuS & Bam & \textbf{Avg.} &
Score & \textbf{Succ.} \\

\midrule
\multicolumn{18}{c}{\textit{\textbf{Teacher Model: Qwen3-4B-Instruct}}} \\
\midrule

GRPO
& 79.3 & 50.0 & 16.1 & 63.2 & 21.7 & 30.0 & 43.4
& 45.4 & 64.1 & 48.2 & 41.6 & 46.1 & 17.7 & 19.8 & 40.4
& 71.4 & 46.1 \\

\midrule
\multicolumn{18}{c}{\textit{\textbf{Distilled Student Model: Qwen3-0.6B}}} \\
\midrule

Vanilla
& 0.0 & 0.0 & 0.0 & 0.0 & 0.0 & 5.0 & 0.8
& 11.9 & 16.4 & 13.3 & 10.5 & 20.7 & 1.3 & 1.6 & 10.8
& 34.8 & 2.3 \\

OPD
& \second{71.4} & 38.5 & 14.8 & \best{50.0} & 4.0 & 0.0 & 29.8
& 7.2 & 9.3 & 12.4 & 6.2 & 4.1 & 0.8 & 2.0 & 6.0
& 3.7 & 0.8 \\

GRPO
& 17.1 & 15.4 & 29.6 & 18.8 & 16.0 & 8.3 & 17.5
& \best{39.6} & \second{54.1} & \best{44.2} & \best{32.8}
& \best{29.5} & \best{9.6} & 11.7 & \second{31.6}
& 54.9 & 10.9 \\

OPSD
& 4.1 & 5.5 & 6.4 & 4.3 & 0.0 & 0.0 & 3.4
& 0.0 & 0.2 & 0.1 & 0.1 & 0.3 & 0.0 & 0.0 & 0.1
& 36.7 & 0.7 \\

TCOD
& 11.4 & 23.1 & 7.4 & 0.0 & 4.0 & 4.2 & 8.3
& 17.9 & 19.0 & 19.6 & 14.4 & 12.5 & 5.3 & 0.0 & 12.7
& 9.0 & 5.5 \\

SDAR
& 50.0 & 33.3 & 40.6 & 30.8 & \second{28.6} & \best{29.2} & 35.4
& 36.6 & 49.4 & 40.4 & 28.5 & \second{28.3} & 6.6
& \second{20.6} & 30.1
& 44.3 & 10.9 \\

Skill-SD
& 44.4 & 33.3 & 28.1 & 23.1 & 14.3 & 12.5 & 26.0
& 38.8 & 53.5 & \second{43.3} & 29.4 & 27.8 & 7.8 & 12.1 & 30.4
& \second{70.6} & \second{49.4} \\

AgentOPSD
& 58.6 & 50.0 & \second{54.8} & 36.8 & 13.0 & 20.0 & 38.9
& 32.8 & 48.1 & 41.5 & 21.1 & 21.4 & 4.1 & 3.2 & 24.6
& 63.3 & 24.2 \\

ATOD
& \best{91.2} & \second{53.3} & 17.4 & \second{42.9}
& 25.0 & 6.7 & \second{39.4}
& 32.8 & 50.3 & 38.1 & 27.3 & 25.4 & 6.7 & 8.9 & 27.1
& 34.8 & 23.4 \\

Ours
& 65.1 & \best{75.0} & \best{66.7} & 33.3
& \best{79.0} & \second{22.2} & \best{56.9}
& \second{39.2} & \best{54.6} & 43.1 & \second{31.8}
& 26.2 & \second{8.6} & \best{23.3} & \best{32.4}
& \best{80.1} & \best{58.6} \\

\midrule
\multicolumn{18}{c}{\textit{\textbf{Distilled Student Model: Qwen3-1.7B}}} \\
\midrule

Vanilla
& 24.1 & 33.3 & 3.2 & 0.0 & 0.0 & 5.0 & 10.9
& 29.4 & 46.9 & 37.0 & 23.5 & 19.6 & 6.4 & 10.5 & 24.8
& 71.4 & 46.1 \\

OPD
& 65.7 & 38.5 & 18.5 & 50.0 & 20.0 & 8.3 & 33.5
& 8.1 & 18.7 & 13.2 & 6.2 & 4.6 & 1.4 & 5.6 & 8.3
& 26.5 & 2.3 \\

GRPO
& 70.0 & 44.4 & 42.8 & 47.4 & 26.3 & 17.4 & 41.4
& 40.0 & \second{58.9} & 43.5 & \second{35.4}
& 30.3 & 12.0 & \second{65.7} & 40.8
& 72.1 & 40.6 \\

OPSD
& 22.2 & 22.2 & 3.1 & 0.0 & 14.3 & 4.2 & 11.0
& 4.2 & 8.3 & 4.6 & 6.6 & 15.3 & 0.7 & 1.2 & 5.8
& 42.1 & 3.9 \\

TCOD
& 42.9 & 38.5 & 3.7 & 12.5 & 0.0 & 12.5 & 18.3
& 23.9 & 35.7 & 25.0 & 19.1 & 18.8 & 1.4 & 0.0 & 17.7
& 44.1 & 5.5 \\

SDAR
& 73.5 & 25.0 & \best{76.9} & 33.3
& 40.0 & \second{36.8} & 47.6
& 39.7 & \second{58.9} & \second{45.3} & \best{35.9}
& \best{35.3} & \second{12.6} & 65.3 & \second{41.9}
& 74.7 & \second{54.4} \\

Skill-SD
& 52.9 & 37.5 & 69.2 & 42.9
& \second{60.0} & \second{36.8} & \second{49.9}
& 39.1 & 57.5 & \best{45.4} & 34.8
& \second{34.1} & 10.7 & 64.1 & 40.8
& \best{81.8} & 53.9 \\

AgentOPSD
& 44.8 & \best{66.7} & 29.0 & \second{63.2}
& 30.4 & 35.0 & 44.8
& \second{41.1} & 55.9 & 43.2 & 33.4 & 28.7 & 10.7 & 62.1 & 39.3
& 65.4 & 36.7 \\

ATOD
& \best{86.2} & \best{66.7} & 35.5 & \best{68.4}
& 8.7 & 10.0 & 45.9
& 32.6 & 46.6 & 32.0 & 23.0 & 19.3 & 7.1 & 18.1 & 25.5
& 43.0 & 35.1 \\

Ours
& \second{82.4} & \second{45.0} & \second{72.7} & 57.1
& \best{70.6} & \best{47.6} & \best{62.6}
& \best{42.1} & \best{59.3} & \best{45.4} & \best{35.9}
& 33.9 & \best{13.1} & \best{68.6} & \best{42.6}
& \second{77.6} & \best{55.7} \\

\bottomrule
\end{tabular}
\end{adjustbox}
\end{table*}

\section{Experiments}


\subsection{Experimental Setup}

\textbf{Benchmarks}. We evaluate our method on two major categories of benchmarks to comprehensively assess the model's performance across tool-use scenarios of varying complexity. The first category consists of classic agent interaction benchmarks, including the text-embodied environment ALFWorld~\citep{shridhar2020alfworld} , SearchQA~\citep{jin2025search} covering single-hop and multi-hop question answering, and the interactive online shopping environment WebShop~\citep{yao2022webshop}. The second category comprises benchmarks specifically designed for complex tool invocation, including the FTRL~\cite{ye2026feedback} dataset, which covers single-step, sequential, and parallel tool-calling scenarios, and the BFCL~\citep{patil2025berkeley} and ToolHop~\citep{ye2025toolhop} dataset, which focus on evaluating long-horizon reasoning and execution capabilities with multi-step tool dependencies. Detailed descriptions are provided in the Appendix~\ref{ap:benchmark}.

\textbf{Baselines.} We compare ToolOPD against three categories of methods: (1) \textbf{Training-free}: Vanilla (the instruction-tuned model without task-specific training). (2) \textbf{Outcome RL and General (Self-) Distillation}: including GRPO~\citep{guo2025deepseek}, ToolRL~\citep{qian2026toolrl}, and MatchTIR~\citep{qu2026matchtir} which optimize outcome rewards, as well as OPD, TCOD~\citep{wang2026tcod}, OPSD~\citep{zhao2026self}, ATOD~\citep{tan2026atod}, and TurnSight~\citep{qu2026turnsight} which employ general distillation without explicit skill construction. (3) \textbf{Skill-Conditioned Self-Distillation}: Skill-SD~\citep{wang2026skill}, SDAR~\citep{lu2026self}, and AgentOPSD~\citep{wang2026agentopsd}, which incorporate skill guidance or trajectory reflection mechanisms. Details of each baseline are provided in the Appendix~\ref{ap:baselines}.

\textbf{Training Details}. All experiments are conducted on 8 A800 GPUs. We adopt Qwen3-0.6B and Qwen3-1.7B as student models, and a Qwen3-4B model trained with GRPO (150-step checkpoint) as the teacher model. In Table~\ref{tab:horizon_results}, the training and optimization setups for ALFWorld, SearchQA, and WebShop are kept consistent with the AgentOPSD baseline. In Table 2, the models are trained exclusively on the FTRL dataset, with the remaining datasets used for out-of-distribution (OOD) generalization evaluation, and their base parameter settings aligned with TurnSight. Detailed hyperparameter configurations are provided in the Appendix~\ref{ap:implementation_detials}.  The evaluation metrics used in this paper are presented in Appendix~\ref{ap:evl_metric}.

\begin{table*}[t]
\centering
\caption{Performance of the teacher and distilled student models on FTRL, BFCL, and ToolHop.}
\label{tab:tool-use_agent_results}
\small
\setlength{\tabcolsep}{3pt}
\renewcommand{\arraystretch}{1.1}
\begin{adjustbox}{max width=\textwidth}
\begin{tabular}{l*{13}{c}}
\toprule
\multirow{2}{*}{Method}
& \multicolumn{4}{c}{FTRL}
& \multicolumn{7}{c}{BFCL}
& \multicolumn{1}{c}{ToolHop}
& \multirow{2}{*}{Avg.}\\
\cmidrule(lr){2-5}
\cmidrule(lr){6-12}
\cmidrule(lr){13-13}
& Solve-F1 & Solve-R & Solve-P & Avg.
& Base & MF & MP & LC & MT.OA & Memory & Avg.
& Acc. & \\
\hline

\multicolumn{14}{c}{\textbf{\textit{Teacher: Qwen3-4B-Instruct}}} \\
\hline
GRPO
& 33.9 & 33.3 & 36.7 & 34.6
& 39.5 & 29.5 & 32.5 & 27.5 & 32.3 & 17.9 & 25.1
& 30.0 & 29.9 \\
\hline

\multicolumn{14}{c}{\textbf{\textit{Student: Qwen3-0.6B}}} \\
\hline
Vanilla
& 9.3 & 8.5 & \second{40.0} & 19.3
& 4.0 & \second{2.0} & \second{3.0} & \second{2.5} & 2.9 & \second{9.3} & \second{6.1}
& 13.0 & 12.8 \\
GRPO
& 13.3 & 13.7 & 25.4 & 17.5
& 3.5 & \second{2.0} & 2.0 & \second{2.5} & 2.5 & 5.8 & 4.2
& \second{17.9} & 13.2 \\
ToolRL
& \best{16.2} & \second{15.8} & 20.4 & 17.5
& 2.5 & 0.0 & 1.5 & 1.5 & 1.4 & 8.4 & 4.9
& 16.9 & 13.1 \\
MatchTIR
& 14.0 & 14.9 & 18.9 & 15.9
& 4.0 & 1.5 & \second{3.0} & \best{3.5} & \second{3.0} & 6.5 & 4.7
& 17.8 & 12.8 \\
TurnSight
& 12.3 & 12.8 & 18.7 & 14.6
& 1.5 & 0.5 & 1.0 & 0.5 & 0.9 & 5.4 & 3.1
& 17.5 & 11.7 \\
SDAR
& \second{16.0} & \best{16.7} & 36.8 & \second{23.2}
& \second{4.5} & 3.0 & 1.5 & 1.5 & 2.6 & 8.4 & 5.5
& 17.7 & \second{15.5} \\
Ours
& \best{16.2} & \second{15.8} & \best{42.7} & \best{24.9}
& \best{5.0} & \best{4.0} & \best{4.5} & \best{3.5} & \best{3.3} & \best{10.0} & \best{7.0}
& \best{18.5} & \best{16.8} \\
\hline

\multicolumn{14}{c}{\textbf{\textit{Student: Qwen3-1.7B}}} \\
\hline
Vanilla
& 19.9 & 22.5 & 21.1 & 21.2
& 15.5 & 11.0 & 12.0 & 7.5 & 11.8 & 9.3 & 10.4
& 20.4 & 17.3 \\
GRPO
& \second{22.6} & 23.6 & \second{25.3} & \second{23.8}
& 18.0 & 15.5 & 12.0 & 11.5 & 14.3 & 9.9 & 12.1
& \second{23.8} & \second{19.9} \\
ToolRL
& 20.4 & 22.3 & 21.0 & 21.2
& 16.0 & 11.0 & 7.0 & 8.0 & 11.1 & \second{11.8} & 11.0
& 19.9 & 17.5 \\
MatchTIR
& 21.6 & \second{24.5} & 22.7 & 23.0
& 13.0 & 14.5 & \second{15.0} & 10.0 & 13.1 & 8.8 & 11.0
& 20.3 & 18.1 \\
TurnSight
& 21.7 & \best{25.6} & 21.3 & 22.9
& \best{24.0} & \second{16.0} & \best{20.0} & \second{13.0} & \best{18.3} & 8.2 & \second{13.2}
& 23.0 & 19.7 \\
SDAR
& 22.4 & \best{25.6} & 22.8 & 23.6
& 18.0 & \best{17.0} & 14.0 & 10.0 & \second{14.9} & 5.8 & 10.3
& 23.1 & 19.0 \\
Ours
& \best{22.7} & 23.9 & \best{28.4} & \best{25.0}
& \second{19.0} & 13.0 & 13.0 & \best{16.0} & 13.4 & \best{12.4} & \best{13.6}
& \best{27.8} & \best{22.2} \\
\bottomrule
\end{tabular}%
\end{adjustbox}
\end{table*}

\subsection{Main Results
}

Tables~\ref{tab:horizon_results} and ~\ref{tab:tool-use_agent_results} present the performance of ToolOPD across various tool-use benchmarks. Overall, ToolOPD significantly outperforms existing reinforcement learning and self-distillation baselines in all evaluated scenarios, validating its effectiveness in addressing the credit assignment problem in long-horizon tool use.

\textbf{Credit Assignment Breakthrough in Long-Horizon Interactions (Table~\ref{tab:horizon_results}).} In long-horizon interaction tasks such as ALFWorld, SearchQA, and WebShop, task success or failure is typically determined by only a few critical tool decisions. Compared to pure reinforcement learning methods (e.g., GRPO), ToolOPD achieves substantial improvements (e.g., the success rate of Qwen3-0.6B on WebShop jumps from 10.9\% to 58.6\%), demonstrating that dense and precise token-level supervision can effectively overcome the credit assignment failure caused by terminal sparse rewards. Furthermore, compared to standard OPD, which implicitly assumes uniform weights (tending to distribute gradients across redundant actions), and AgentOPSD, which lacks outcome filtering, ToolOPD precisely quantifies and amplifies the weights of critical actions that substantially contribute to the final outcome through the outcome-asymmetric mechanism ($H_{i,t}$), fundamentally overcoming the dilution of core decision signals.

\textbf{Complex Tool Dependencies and Generalization Capabilities (Table~\ref{tab:tool-use_agent_results}).} Across the FTRL, BFCL, and ToolHop benchmarks, ToolOPD not only achieves the highest average performance on the FTRL training set but also demonstrates strong generalization capabilities on out-of-distribution (OOD) test sets such as ToolHop and BFCL. Particularly in the ToolHop task, which features extremely long interaction turns and a large number of ineffective retrievals, ToolOPD significantly outperforms the second-best baseline, SDAR (improving the accuracy of Qwen3-1.7B from 19.0\% to 22.2\%). This is attributed to ToolOPD's hindsight skill induction mechanism, which compresses macroscopic trajectory outcomes into microscopic strategy priors, combined with asymmetric relevance to truncate gradients for ineffective actions. This ensures that the model can precisely pinpoint core API calls within long-horizon trajectories, effectively enhancing reasoning and execution capabilities under complex state transitions.

\begin{table}[t]
\centering
\caption{Ablation study on the ALFWorld benchmark. We progressively add components to GRPO: standard action-level distillation (Vanilla OPD), hindsight skill conditioning (Skill-OPD), and the full ToolOPD. The best results in each column are highlighted in bold.}
\label{tab:ablation}
\small
\setlength{\tabcolsep}{1pt}
\renewcommand{\arraystretch}{1}
\begin{tabular}{lccccccc}
\toprule
\textbf{Method}
& \textbf{Pick}
& \textbf{Look}
& \textbf{Clean}
& \textbf{Heat}
& \textbf{Cool}
& \textbf{Pick2}
& \textbf{Scc.} \\
\midrule
GRPO
& 17.1 & 15.4 & 29.6 & 18.8 & 16.0 & 8.3 & 14.8 \\
GRPO + Vanilla OPD
& 55.6 & 66.7 & 62.5 & 30.7 & \textbf{92.9} & 25.0 & 53.9 \\
GRPO + Skill-OPD
& \textbf{82.4} & 50.0 & 46.2 & 19.1 & 60.0 & \textbf{52.6} & 54.7 \\
\textbf{ToolOPD (Ours)}
& 65.1 & \textbf{75.0} & \textbf{66.7} & \textbf{33.3} & 79.0 & 22.2 & \textbf{59.4} \\
\bottomrule
\end{tabular}%
\end{table}

\subsection{Ablation Study}
\label{sec:ablation}

To investigate the contribution of each component in ToolOPD to long-horizon credit assignment, we conduct a progressive ablation study on the ALFWorld benchmark, as reported in Table~\ref{tab:ablation}. GRPO, which relies solely on sparse trajectory-level rewards, achieves an average success rate of only 14.8\%, highlighting the limitation of sequence-level supervision in identifying pivotal decisions during long-horizon tool interactions. Incorporating standard action-level distillation substantially improves the average success rate to 53.9\%, demonstrating that dense teacher supervision over action tokens provides effective local learning signals. However, Vanilla OPD assigns relatively uniform distillation weights across interaction steps and therefore cannot adequately distinguish pivotal decisions from redundant actions. Further introducing hindsight skill conditioning yields only a marginal improvement to 54.7\%, with performance degradation on several task categories. This result suggests that skill conditioning alone does not guarantee consistent gains, as the model must additionally determine whether the induced skill is aligned with the trajectory outcome and whether the corresponding teacher signal is reliable.

The complete ToolOPD combines outcome-asymmetric skill relevance $H_{i,t}$ with teacher confidence and experience-based reliability to emphasize actions that are more relevant to the final outcome. It achieves the best average success rate of 59.4\%, outperforming Vanilla OPD and Skill-OPD by 5.5 and 4.7 percentage points, respectively. The largest gains over Skill-OPD occur on Look, Clean, and Heat, with improvements of 25.0, 20.5, and 14.3 points. These results show that outcome-aware relevance and reliability calibration are critical for robust credit assignment in long-horizon tool interactions.


\begin{figure*}[t]
    \centering
    \begin{subfigure}[t]{0.24\textwidth}
        \centering
        \includegraphics[width=\textwidth]{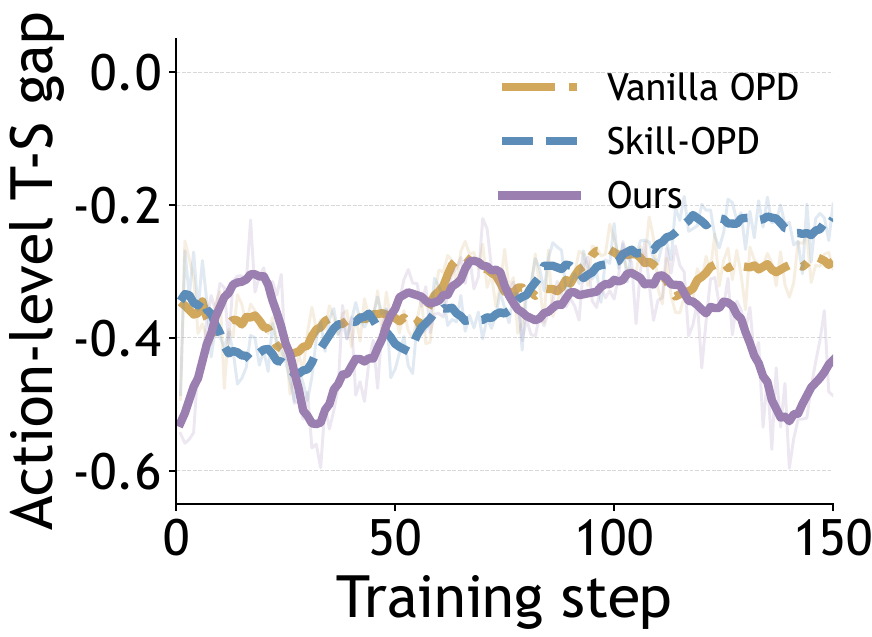}
        \caption{Teacher-student gap.}
        \label{fig:alfworld-gap}
    \end{subfigure}
    \hfill
    \begin{subfigure}[t]{0.24\textwidth}
        \centering
        \includegraphics[width=\textwidth]{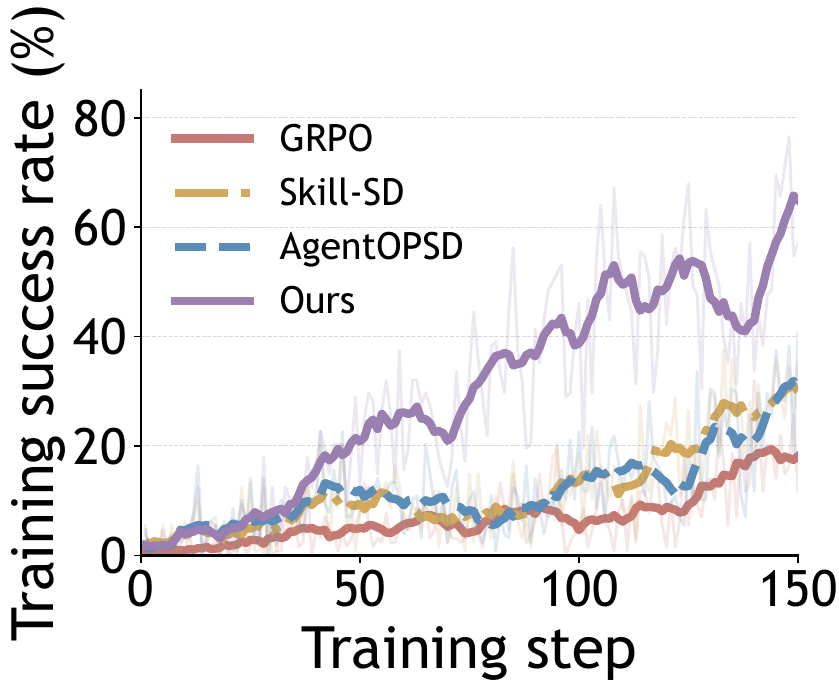}
        \caption{ALFWorld success rate.}
        \label{fig:alfworld-success}
    \end{subfigure}
    \begin{subfigure}[t]{0.24\textwidth}
        \centering
        \includegraphics[width=\textwidth]{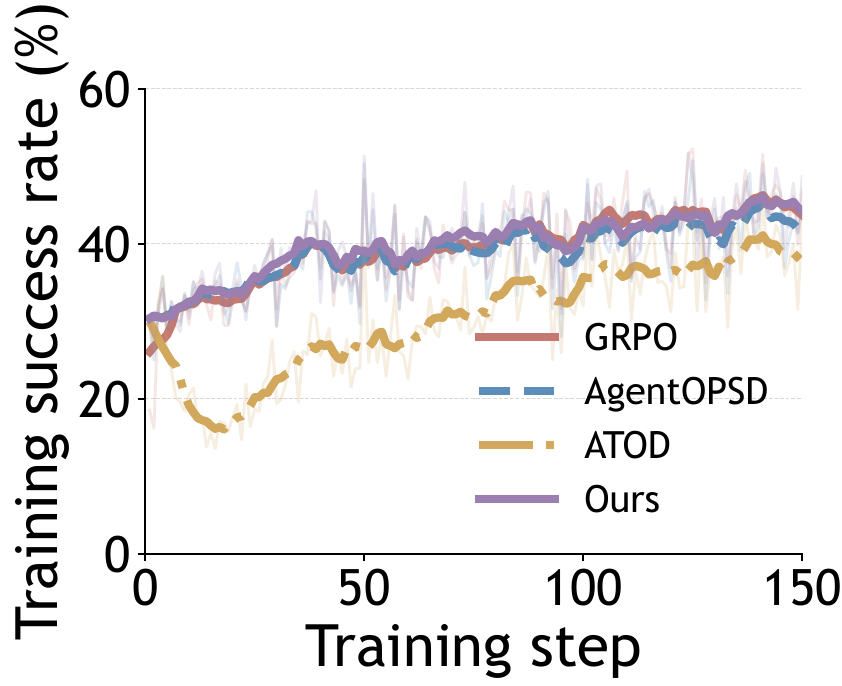}
        \caption{SearchQA success rate.}
        \label{fig:searchqa-success}
    \end{subfigure}
    \hfill
    \begin{subfigure}[t]{0.24\textwidth}
        \centering
        \includegraphics[width=\textwidth]{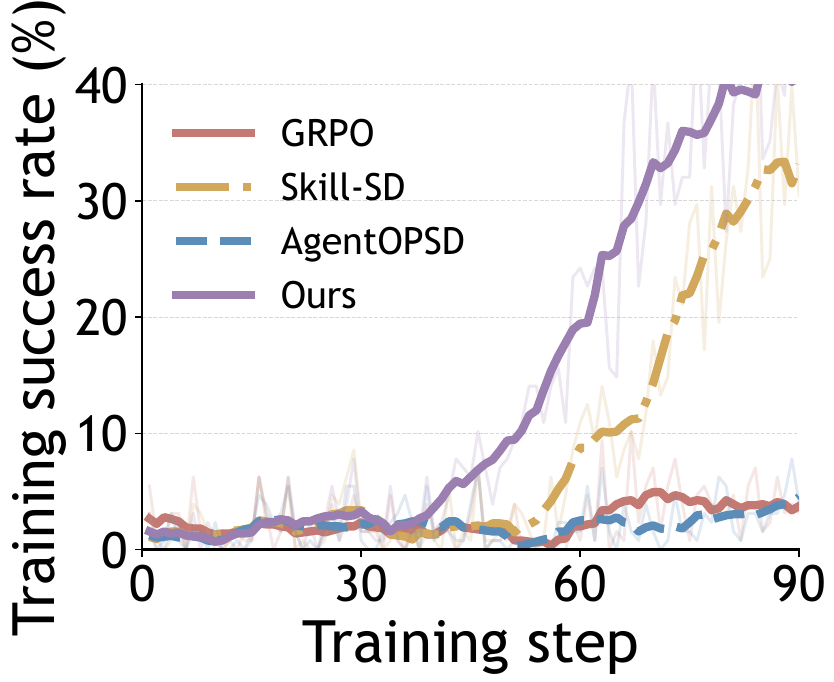}
        \caption{WebShop success rate.}
        \label{fig:webshop-success}
    \end{subfigure}

    \caption{Training dynamics across different tasks.}
    \label{fig:training-dynamics}
\end{figure*}


\begin{figure*}[t]
    \centering
    \begin{subfigure}[t]{0.24\textwidth}
        \centering
        \includegraphics[width=\textwidth]{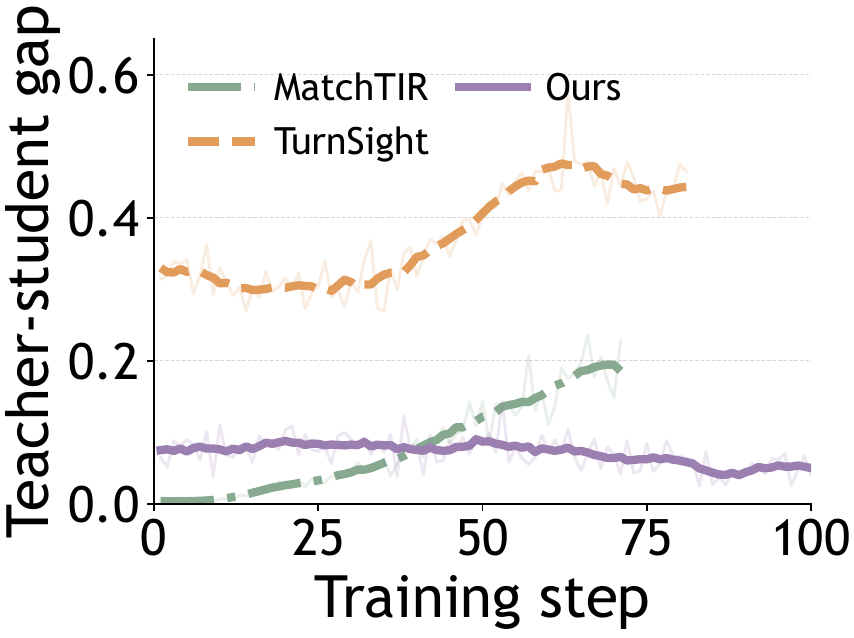}
        \caption{Teacher-student gap.}
        \label{fig:ftrl-teacher-student-gap}
    \end{subfigure}
    \hfill
    \begin{subfigure}[t]{0.24\textwidth}
        \centering
        \includegraphics[width=\textwidth]{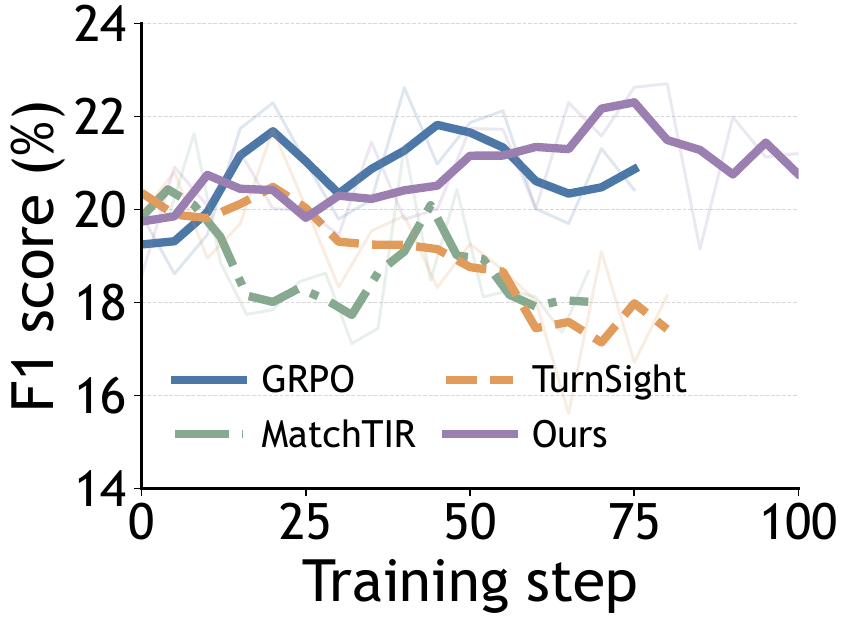}
        \caption{FTRL F1 score.}
        \label{fig:ftrl-f1-score}
    \end{subfigure}
    \hfill
    \begin{subfigure}[t]{0.24\textwidth}
        \centering
        \includegraphics[width=\textwidth]{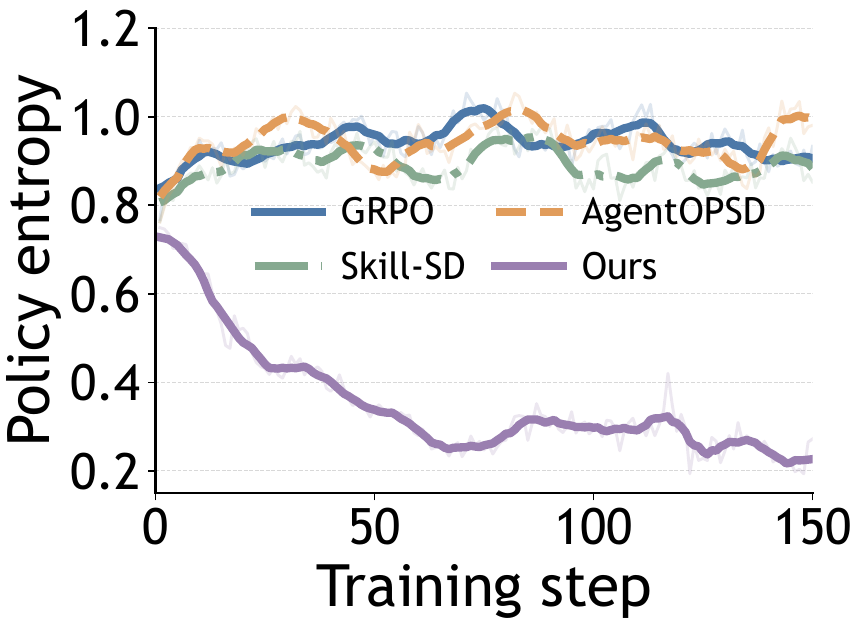}
        \caption{Policy entropy.}
        \label{fig:alfworld-policy-entropy}
    \end{subfigure}
    \hfill
    \begin{subfigure}[t]{0.24\textwidth}
        \centering
        \includegraphics[width=\textwidth]{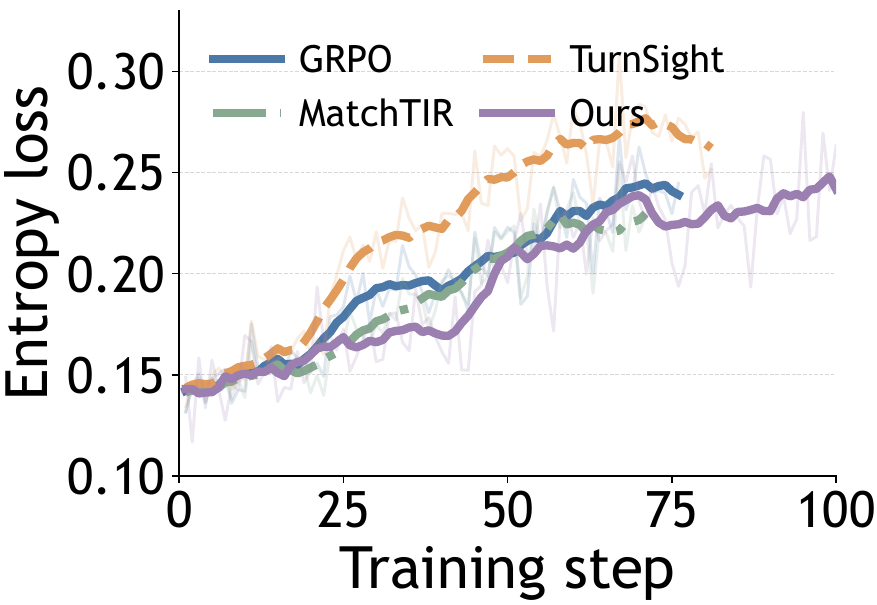}
        \caption{FTRL entropy loss.}
        \label{fig:ftrl-entropy-loss}
    \end{subfigure}

    \caption{Training dynamics of the model across ALFWorld (c) and FTRL (a, b, and d).}
    \label{fig:training-dynamic-v2}
\end{figure*}

\subsection{Training Dynamics}
Figure~\ref{fig:training-dynamics} presents the training dynamics of different methods across multiple tool-interaction benchmarks. ToolOPD achieves faster and more sustained performance improvements on ALFWorld and WebShop, establishing a clear advantage over the baselines in the later stages of training. Although the differences among methods are smaller on SearchQA, ToolOPD still attains the best final performance. These results suggest that its benefits are not specific to a particular environment, but generalize across tasks with different interaction horizons and reward structures. Meanwhile, the teacher--student gap in Figure~3(a) is not monotonically minimized toward zero, but instead changes dynamically throughout training. This behavior indicates that ToolOPD does not uniformly imitate all teacher signals; rather, it adaptively adjusts the strength of distillation according to the current policy state and trajectory outcome.

Fig.~\ref{fig:training-dynamic-v2} further characterizes the optimization behavior underlying these performance gains. On FTRL, ToolOPD maintains a consistently small and stable teacher--student discrepancy, whereas the discrepancies of MatchTIR and TurnSight progressively increase during training. Correspondingly, ToolOPD improves its F1 score more rapidly and reaches a higher level in the middle and later stages. On ALFWorld, the policy entropy of ToolOPD decreases substantially while its success rate continues to improve, suggesting that the policy gradually shifts from broad exploration toward more concentrated tool decisions rather than merely suffering from performance degradation. The entropy dynamics on FTRL exhibit a similar pattern: ToolOPD shows a smoother increase in entropy and avoids the pronounced growth in uncertainty observed for TurnSight. Overall, these results indicate that outcome-aware dynamic distillation enables ToolOPD to control teacher--student discrepancy while progressively focusing optimization on critical actions, leading to a more stable and efficient policy learning process.

\begin{wrapfigure}{r}{0.48\textwidth}
    \centering
    \setlength{\abovecaptionskip}{5pt}
    \setlength{\belowcaptionskip}{-1pt}
    \vspace{-0.5em}
    \includegraphics[width=0.44\textwidth]{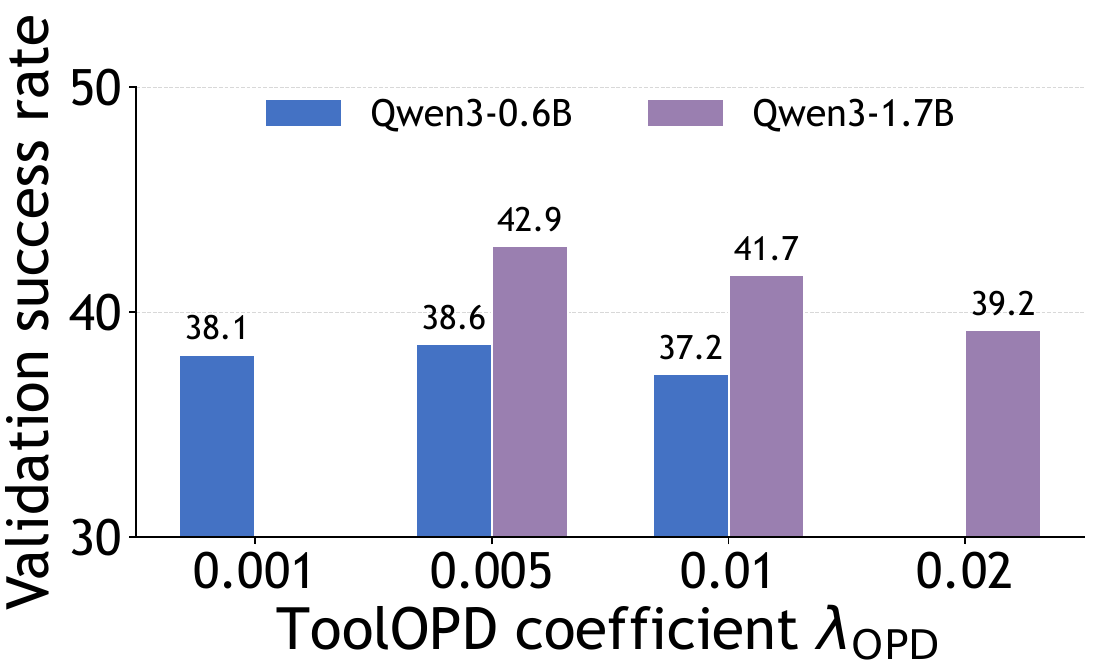}
    \caption{Hyperparameter sensitivity analysis.}
    \label{fig:toolopd-coefficient-sensitivity}
    \vspace{-0.5em}
\end{wrapfigure}
\subsection{Hyperparameter Analysis of ToolOPD}
To assess the effect of the ToolOPD coefficient $\lambda_{\mathrm{OPD}}$, we conduct a sensitivity analysis on SearchQA using both Qwen3-0.6B and Qwen3-1.7B. As shown in Fig.~\ref{fig:toolopd-coefficient-sensitivity}, both models exhibit a similar inverted-U-shaped trend: validation success first improves as $\lambda_{\mathrm{OPD}}$ increases to $0.005$, and then declines when the coefficient is further increased. Specifically, the 0.6B model achieves its best success rate of 38.58\% at $\lambda_{\mathrm{OPD}}=0.005$, outperforming the settings of $0.001$ (38.11\%) and $0.01$ (37.24\%). The 1.7B model follows the same pattern, reaching 42.94\% at $0.005$, while performance decreases to 41.66\% and 39.20\% at $0.01$ and $0.02$, respectively. These results suggest that a moderate action-level distillation weight provides effective dense supervision: a coefficient that is too small is insufficient to leverage ToolOPD for credit assignment, whereas an overly large coefficient can cause teacher supervision to dominate policy optimization. Based on this trend, we use $\lambda_{\mathrm{OPD}}=0.005$ as the default setting for SearchQA.

\begin{wrapfigure}{r}{0.5\textwidth}
    \centering
    \vspace{-0.5em}
    \centering
\setlength{\abovecaptionskip}{3pt}
\setlength{\belowcaptionskip}{-1pt}
\begin{subfigure}[t]{0.49\linewidth}
    \centering
    \includegraphics[width=\linewidth]{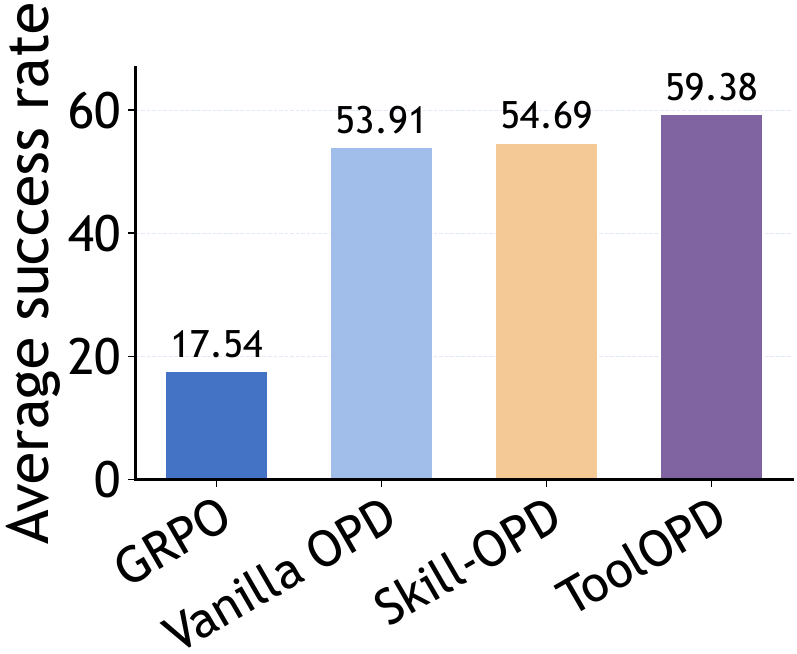}
    \label{fig:toolopd-credit-ablation-performance}
\end{subfigure}
\begin{subfigure}[t]{0.49\linewidth}
    \centering
    \includegraphics[width=\linewidth]{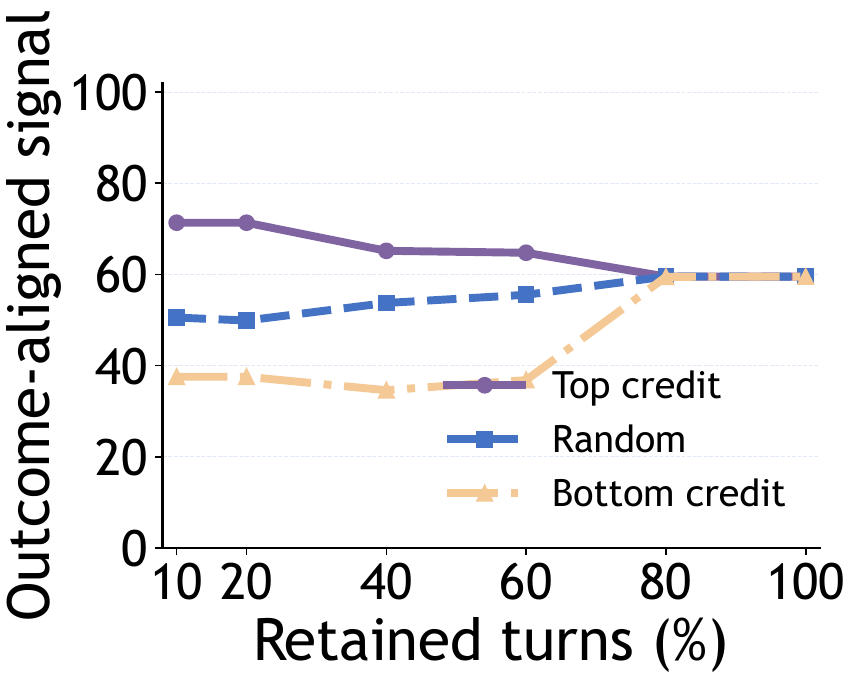}
    \label{fig:toolopd-topk-signal-quality}
\end{subfigure}
\caption{Further analysis of Tool-OPD. \textbf{(Left)} Performance under different credit-assignment ablation settings. \textbf{(Right)} Quality of training signals with different top-$k$ selections.}
\label{fig:toolopd-further-analysis}
 \vspace{-0.5em}
\end{wrapfigure}

\subsection{Further Analysis of Credit Assignment}
To further examine the credit assignment capability of ToolOPD, we analyze both performance gains and supervision quality. As shown in Fig.~\ref{fig:toolopd-further-analysis}(a), ToolOPD achieves an average success rate of 59.38\%, outperforming Vanilla OPD and Skill-OPD by 5.47 and 4.69 percentage points, respectively. This result indicates that dense distillation or hindsight skills alone are insufficient to achieve optimal performance. Fig.~\ref{fig:toolopd-further-analysis}(b) further explains the source of this improvement. When considering only the top 10\%--20\% highest-credit turns, the proportion of outcome-aligned signals reaches approximately 71.3\%, substantially higher than random turns (approximately 50.0\%) and bottom-credit turns (approximately 37.6\%). As the retained proportion increases, the differences gradually diminish and eventually converge. These results demonstrate that ToolOPD effectively ranks turns according to supervision quality and prioritizes distillation signals associated with tool decisions that are more likely to affect the final outcome. Taken together, Fig.~~\ref{fig:toolopd-further-analysis} (a) and (b) suggest that the performance gains of ToolOPD primarily arise from fine-grained, outcome-aware credit assignment, rather than merely introducing additional teacher supervision.


\section{Related Work}

\subsection{On-policy Distillation}
On-policy distillation (OPD)~\citep{song2026survey} aligns the student and teacher policies by matching the student policy to the teacher distribution over its self-generated trajectories. By utilizing objectives such as token-level KL divergence, OPD effectively mitigates the state distribution shift inherent in offline distillation.
As OPD is extended to long-horizon agent tasks, research has transitioned from uniform token-level matching to step- or turn-level supervision. In Toolintegrated scenarios, existing approaches primarily operate along two dimensions. First, they leverage dynamic skills or hindsight information from completed trajectories to guide self-distillation (including Skill-SD~\citep{wang2026skill}, SEED~\citep{wu2026seed}, and TurnSight~\cite{qu2026turnsight}). Second, they dynamically adjust the distillation weights across interaction turns based on factors including teacher-student divergence, training budgets, and interaction difficulty (including SOD~\citep{zhong2026sod}, TurnOPD~\citep{zhou2026turnopd}, AgentOPSD~\citep{wang2026agentopsd}, and HIDIFFTIR~\citep{guo2026hidifftir}). These studies demonstrate that online distillation for Toolusing agents must model multi-turn interaction structures, tool invocation difficulty, and hindsight information, rather than relying on uniform token-level supervision.

In contrast to existing methods, this paper focuses on the credit assignment problem for small models executing tool invocations in long-horizon tasks on resource-constrained terminal devices. We propose ToolOPD, which moves beyond the reliance on local training difficulty or teacher-student divergence by directly evaluating the true contribution of tool decisions based on final task outcomes. The core of ToolOPD lies in performing non-uniform credit assignment for distinct tool decisions and dynamically adjusting the distillation intensity of action tokens accordingly, thereby maximizing the distillation efficacy of small models under resource constraints.

\subsection{Credit Assignment in Long-Horizon On-Policy Distillation}
In long-horizon tool use tasks~\citep{feng2026retool,jiang2025verltool,qian2026toolrl}, standard OPD typically performs uniform token-level distribution matching across all visited states. However, because different interaction turns contribute unequally to task success, applying supervision with uniform intensity is highly prone to introducing noise.
Recent studies have introduced fine-grained credit assignment into long-horizon interactions and policy distillation. Some methods evaluate intermediate action values by quantifying information gain, leveraging hindsight information, or decomposing trajectory-level outcomes (e.g., $\Delta$Belief-RL~\citep{auzina2026intrinsic}, HCAPO~\citep{tan2026hindsight}, and PBSD~\citep{tian2026pbsd}). Others identify critical action spans and adjust local advantages from the perspectives of supervision reliability, failure correlation, or policy comparison (e.g., Prune-OPD~\citep{yang2026prune}, HINT-SD~\citep{yeo2026hint}, UCOB~\citep{tu2026ucob}, and SGCD~\citep{ding2026keep}). Overall, existing research has transitioned from uniform token-level matching to credit-aware selective supervision.

In this work, ToolOPD locates critical tool decisions by quantifying the shift in the teacher's action distribution conditioned on hindsight skills. It then performs multi-dimensional credit calibration by integrating the final outcome, trajectory consistency, and teacher confidence. Ultimately, the calibrated turn-level credit is directly mapped to the action tokens actually sampled by the student, achieving non-uniform on-policy distillation oriented towards long-horizon task outcomes.


\section{Conclusion}

We introduce ToolOPD, an outcome-aware on-policy distillation framework that addresses credit assignment in long-horizon tool-use tasks. Our empirical analysis shows that the uniform supervision of standard OPD fails to distinguish the contributions of different tool decisions and may introduce teacher signals that conflict with the final outcome. To address this limitation, ToolOPD induces hindsight skills from complete trajectories and their outcomes, and dynamically assigns credit to action tokens through outcome-asymmetric skill relevance. This preserves supervision consistent with the task outcome while suppressing irrelevant or conflicting signals. Experiments across multiple long-horizon tool-interaction benchmarks and model scales demonstrate that ToolOPD consistently outperforms reinforcement-learning and on-policy-distillation baselines while exhibiting more stable optimization over extended interactions.

\clearpage
\bibliography{iclr2026_conference}
\bibliographystyle{iclr2026_conference}

\clearpage
\appendix
\startcontents[appendix]

\section*{Appendix}
\addcontentsline{toc}{section}{Appendix}

\begin{center}
    \large\bfseries Contents
\end{center}

\printcontents[appendix]{l}{1}{
    \setcounter{tocdepth}{2}
}
\clearpage

\section{Benchmarks}
\label{ap:benchmark}
\begin{itemize}[left=0pt]
\item \textbf{ALFWorld} is an interactive benchmark that aligns abstract textual reasoning with concrete embodied environments, requiring agents to translate high-level text policies into multi-step navigation and object manipulation actions in simulated household settings, thereby evaluating cross-modal grounding and long-horizon planning.

\item \textbf{SearchQA} is a large-scale open-domain question answering dataset comprising over 140,000 question-answer pairs augmented with real-world search engine contexts, designed to assess model robustness in precise machine reading comprehension and information extraction amidst highly redundant and noisy retrieved snippets.

\item \textbf{WebShop} provides a simulated e-commerce environment grounded in real-world web interactions, featuring over 1.18 million real products and 12,000+ crowdsourced natural language instructions. It requires agents to perform sequential decision-making to search, filter, and purchase target items, systematically evaluating compositional instruction understanding, dynamic query reformulation, and long-horizon multi-step interaction.

\item \textbf{FTRL} is a foundational evaluation dataset focused on tool-use capabilities, comprising over 2,000 automatically constructed tool interaction scenarios with verifiable execution environments. It typically serves as a basis for in-domain training and evaluation, aiming to systematically measure a model's tool-calling accuracy and sequential decision-making efficacy within standardized environments.
    
\item \textbf{BFCL (Berkeley Function Calling Leaderboard)}, proposed by Patil et al., is a comprehensive function-calling evaluation benchmark. Its core dataset encompasses thousands of question-function-answer pairs spanning multiple programming languages and interfaces (e.g., Python, Java, JavaScript, and REST APIs). Through iterative updates, BFCL has expanded to include complex test cases involving multi-turn dialogues, multi-step dependencies, parallel function calls, and real-time dynamic scenarios, primarily assessing a model's instruction-following capability and dynamic decision-making robustness in real-world API interactions.
    
\item \textbf{ToolHop} is a query-driven benchmark specifically designed to evaluate the multi-hop tool use capabilities of large language models. The dataset consists of 995 complex user queries and 3,912 associated locally executable tools. By constructing toolchains with strong interdependencies, it rigorously examines a model's comprehensive proficiency in long-horizon logical reasoning, dynamic query reformulation, and multi-step tool orchestration.
\end{itemize}

\section{Baselines}
\label{ap:baselines}

\subsection{Outcome RL and General (Self-) Distillation}
\begin{itemize}[left=0pt]
    \item\textbf{GRPO}~\citep{guo2025deepseek} is a reinforcement learning method that eliminates the need for an explicit value model. It samples multiple responses for the same input and estimates advantages based on their relative rewards within the group. Its supervision mainly comes from the final task outcome, such that all decisions along a trajectory typically share the same outcome-level reward.

    \item \textbf{ToolRL}~\citep{qian2026toolrl} is a reinforcement learning framework designed for tool-use tasks. It trains tool-using agents with reward signals associated with task outcomes and tool execution, aiming to improve tool selection and invocation capabilities in complex and unseen tool-use scenarios.

    \item \textbf{MatchTIR}~\citep{qu2026matchtir} addresses the coarse-grained credit assignment problem of trajectory-level rewards in multi-turn tool-integrated reasoning. It matches generated tool-use trajectories with reference trajectories via bipartite matching to construct turn-level rewards, and combines trajectory-level and turn-level advantages for more fine-grained policy optimization.

    \item \textbf{OPD} lets the student model generate trajectories under its current policy and queries a teacher model to provide token-level distribution supervision on states visited by the student. Compared with offline distillation, OPD reduces the train--test distribution mismatch by directly training on the student's on-policy state distribution.

    \item \textbf{TCOD}~\citep{wang2026tcod} focuses on trajectory-level KL instability caused by error accumulation in multi-turn on-policy distillation. It adopts a temporal curriculum that gradually increases the interaction horizon from short trajectories to full-length trajectories, enabling the student to progressively acquire long-horizon interaction capabilities while improving training stability.

    \item \textbf{OPSD}~\citep{zhao2026self} is an on-policy self-distillation approach that removes the dependence on a separate large teacher model. The same model acts as both student and teacher under different contextual conditions: the student generates on-policy trajectories, while a privileged teacher view with additional task information provides token-level supervision.

    \item \textbf{ATOD}~\cite{tan2026atod} is a turn-aware on-policy distillation method for long-horizon multi-turn agents. It combines OPD with GRPO through an annealed training scheme and dynamically adjusts the distillation strength at each turn according to teacher--student disagreement and uncertainty, thereby allocating stronger supervision to interaction steps that remain difficult for the student.

    \item \textbf{TurnSight}~\citep{qu2026turnsight} introduces turn-level hindsight self-distillation for tool-integrated reasoning. It constructs a hindsight teacher using tool execution outcomes and measures teacher--student distribution discrepancies at individual turns. Multi-step lookahead and group-wise normalization are further used to assign differentiated supervision weights across tool-interaction turns.
\end{itemize}
\subsection{Skill-Conditioned Self-Distillation}
\begin{itemize}[left=0pt]
    \item \textbf{Skill-SD}~\citep{wang2026skill} is a skill-conditioned self-distillation framework for multi-turn LLM agents. It summarizes completed agent trajectories into compact natural-language skills describing successful behaviors, common mistakes, and reusable workflows, and uses these skills as dynamic privileged information for the teacher branch. The student is then trained to internalize such skill guidance through on-policy distillation, together with an importance-weighted reverse-KL objective and dynamic teacher synchronization for stable training.

    \item \textbf{SDAR}~\citep{lu2026self} integrates on-policy self-distillation into agentic reinforcement learning as a gated auxiliary objective, while retaining RL as the primary optimization signal. It derives token-level teacher--student discrepancy signals and maps them into adaptive gates, strengthening distillation on teacher-endorsed tokens while attenuating unreliable negative guidance, thereby improving training stability in long-horizon multi-turn tasks.

    \item \textbf{AgentOPSD}~\cite{wang2026agentopsd} introduces recursive self-distillation for fine-grained credit assignment in long-horizon agentic reinforcement learning. It aggregates token-level teacher--student log-probability gaps into turn-level evidence and recursively updates a Bayesian belief state across interaction turns, using the resulting belief revisions to estimate the importance of individual decisions and reweight policy optimization without introducing an additional critic or extra rollouts.
\end{itemize}

\section{Implementation Details}
\label{ap:implementation_detials}
For the long-horizon tasks reported in Table~\ref{tab:horizon_results}, we use GRPO as the base policy optimization algorithm and jointly optimize it with the ToolOPD auxiliary objective. ToolOPD operates exclusively on tool-action tokens. For each completed trajectory, the teacher generates an outcome-conditioned hindsight skill, whose effect on the teacher's action probabilities is converted into fine-grained credit for reweighting the distillation loss. We set the hindsight scaling factor to $\beta_h=5.0$ and the replay calibration factor to $\alpha=1.0$, while linearly increasing the evidence weight $\lambda_e$ from $0.2$ to $0.5$ during training. The ToolOPD loss coefficient $\lambda_{\mathrm{OPD}}$ is selected from $0.001$ to $0.01$ according to the task and student model size, and the log-probability ratio is clipped at $5.0$ for training stability.

We use Qwen3-0.6B or Qwen3-1.7B as the student policy and a GRPO-trained Qwen3-4B as the frozen teacher. The same teacher performs both hindsight skill induction and action-level supervision, avoiding the need for an additional skill-generation model or external API. During training, four GPUs are allocated to the student policy and four to the independent teacher pool. The environment-specific training configurations are summarized in Table~\ref{tab:toolopd_hyperparameters}.

For tool-use benchmark, we augment vanilla GRPO with a ToolOPD auxiliary loss applied only to tool-action tokens. We use Qwen3-0.6B as the student and a GRPO-trained Qwen3-4B as the frozen teacher. We set $\beta_h=10.0$, $\alpha=1.0$, $\lambda_{\mathrm{OPD}}=0.03$, and $\lambda_e=0$, while clipping the teacher--student log-ratio at $3.0$. Training uses eight rollouts per prompt, a batch size of $32$, and a learning rate of $2\times10^{-7}$.
\begin{table}[t]
    \centering
    \caption{Core ToolOPD training configurations across the three tool-use benchmarks.}
    \label{tab:toolopd_hyperparameters}
    \small
    \setlength{\tabcolsep}{5.5pt}
    \renewcommand{\arraystretch}{1.08}
    \begin{tabular}{lccc}
        \toprule
        \textbf{Hyperparameter}
        & \textbf{ALFWorld}
        & \textbf{WebShop}
        & \textbf{SearchQA} \\
        \midrule
        Training steps
        & 150 & 150 & 150 \\
        Training batch size
        & 16 & 16 & 128 \\
        Rollout group size $G$
        & 8 & 8 & 8 \\
        Maximum prompt length
        & 2048 & 4096 & 4096 \\
        Maximum response length
        & 512 & 512 & 512 \\
        Maximum interaction turns
        & 50 & 15 & 4 \\
        Train/validation temperature
        & 1.0 / 0.4 & 1.0 / 0.4 & 1.0 / 0.4 \\
        GPUs (student / teacher)
        & 4 / 4 & 4 / 4 & 4 / 4 \\
        Rollout tensor-parallel size
        & 2 & 2 & 2 \\
        \bottomrule
    \end{tabular}
\end{table}

\begin{table}[t]
    \centering
    \caption{Core training hyperparameters for ToolOPD on the tool-use benchmarks.}
    \label{tab:turnsight_toolopd_hyperparameters}
    \small
    \setlength{\tabcolsep}{8pt}
    \renewcommand{\arraystretch}{1.05}
    \begin{tabular}{lc}
        \toprule
        \textbf{Hyperparameter} & \textbf{Value} \\
        \midrule
        Student model
             & Qwen3-0.6B / Qwen3-1.7B \\
        Teacher model
            & GRPO-trained Qwen3-4B \\
        Policy optimization algorithm
            & Vanilla GRPO \\
        Group size $G$ (rollouts per task)
            & 8 \\
        Training batch size
            & 32 \\
        PPO mini-batch size
            & 32 \\
        Optimization epochs per batch
            & 1 \\
        Learning rate
            & $2\times10^{-7}$ \\
        Training temperature
            & 1.0 \\
        Maximum prompt length
            & 10,000 tokens \\
        Maximum response length
            & 13,000 tokens \\
        Maximum tokens per GPU
            & 23,000 \\
        Tensor parallel size
            & 1 \\
        KL-loss coefficient
            & 0.0 \\
        Entropy coefficient
            & 0.0 \\
        \midrule
        Hindsight scaling factor $\beta_h$
            & 10.0 \\
        Replay calibration factor $\alpha$
            & 1.0 \\
        Evidence weight $\lambda_e$
            & 0.0 \\
        ToolOPD loss coefficient $\lambda_{\mathrm{OPD}}$
            & 0.03 \\
        Maximum log-ratio threshold
            & 3.0 \\
        Hindsight generation length
            & 96 tokens \\
        Teacher maximum prompt length
            & 14,000 tokens \\
        Hindsight maximum prompt length
            & 14,000 tokens \\
        ToolOPD supervision scope
            & Tool-action tokens only \\
        \bottomrule
    \end{tabular}
\end{table}

\begin{figure*}[t]
    \centering
    \begin{subfigure}[t]{0.24\textwidth}
        \centering
        \includegraphics[width=\textwidth]{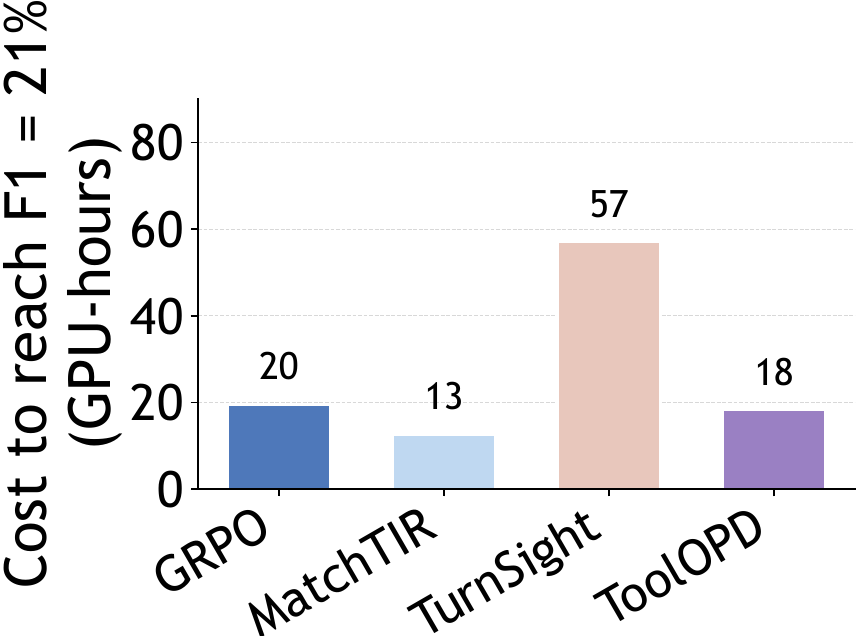}
        \caption{Cost to reach F1 21\%.}
    \end{subfigure}
    \hfill
    \begin{subfigure}[t]{0.24\textwidth}
        \centering
        \includegraphics[width=\textwidth]{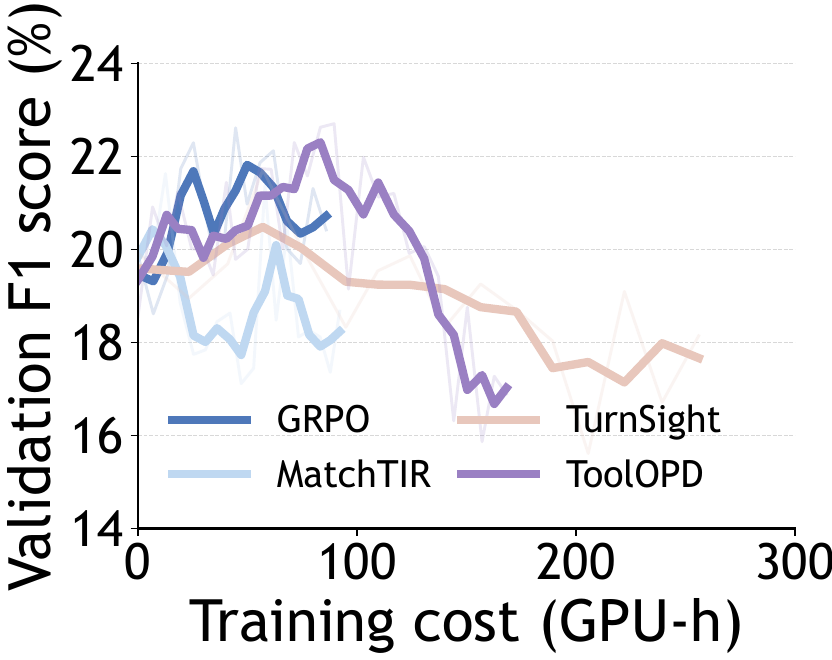}
        \caption{F1 versus GPU hours.}
        \label{fig:ftrl-f1-gpu-hours}
    \end{subfigure}
    \hfill
    \begin{subfigure}[t]{0.24\textwidth}
        \centering
        \includegraphics[width=\textwidth]{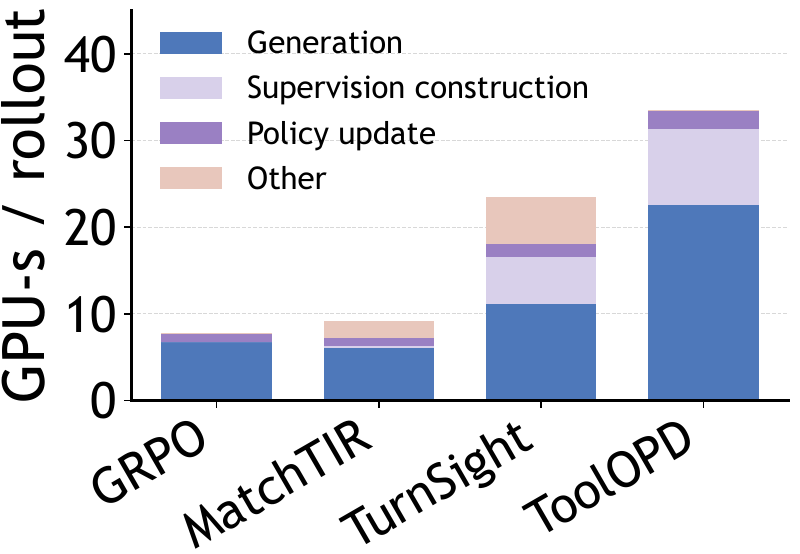}
        \caption{Rollout cost breakdown.}
    \end{subfigure}
    \hfill
    \begin{subfigure}[t]{0.24\textwidth}
        \centering
        \includegraphics[width=\textwidth]{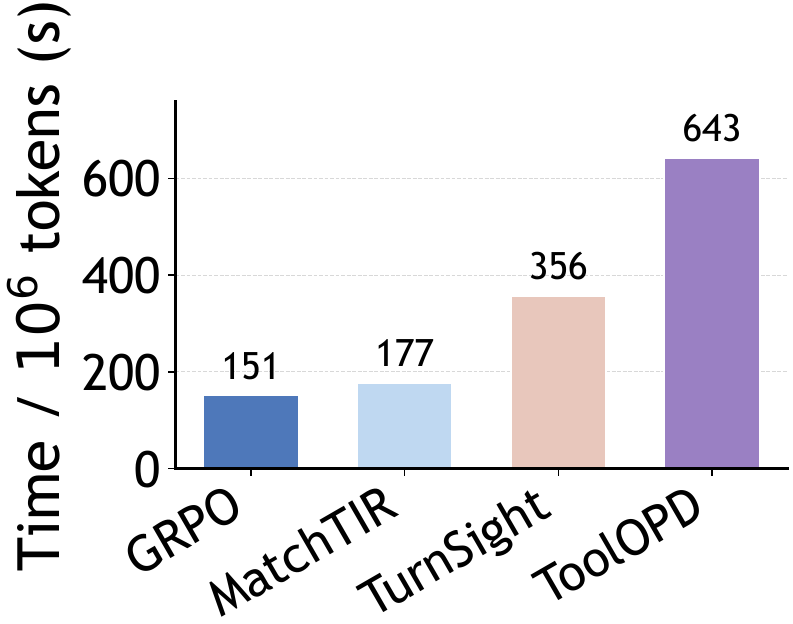}
        \caption{Time per million tokens.}
    \end{subfigure}

    \caption{Computational overhead analysis on FTRL.}
    \label{fig:computational-overhead}
\end{figure*}
\section{Computational Overhead Analysis}
To quantify the additional computational cost of ToolOPD, we use a matched eight-GPU setup on FTRL and exclude evaluation and checkpointing time from all measurements. As shown in Fig.~\ref{fig:computational-overhead}(d), ToolOPD does not improve raw training throughput: it requires 643 seconds per million training tokens, compared with 151, 177, and 356 seconds for GRPO, MatchTIR, and TurnSight, respectively. The per-rollout cost breakdown in Fig.~\ref{fig:computational-overhead}(c) further shows that ToolOPD is dominated by rollout generation (22.5 GPU-s) and outcome-aware supervision construction (8.8 GPU-s), whereas policy updates account for only a small fraction of the total cost (2.1 GPU-s). These results indicate that the additional overhead primarily arises from teacher-driven hindsight skill inference and action-level supervision construction, rather than standard policy optimization.

Despite this overhead, the performance--cost curve in Fig.~\ref{fig:computational-overhead}(b) reveals a favorable optimization trade-off in some regimes. In particular, Fig.~\ref{fig:computational-overhead}(a) shows that ToolOPD reaches an F1 score of 21\% after 18.2 GPU-hours, requiring less computation than GRPO (19.5 GPU-hours) and substantially less than TurnSight (57.0 GPU-hours), although it remains more expensive than MatchTIR (12.5 GPU-hours). Therefore, ToolOPD should not be viewed as a method for improving raw computational efficiency; instead, it trades additional computation for more targeted credit assignment, allowing it to reach a comparable performance level with lower cost than TurnSight.

\section{Algorithm}
\label{ap:algorithm}
\begin{algorithm}[H]
\caption{Outcome-Aware On-Policy Distillation for ToolOPD}

\begin{algorithmic}[1]
\REQUIRE Student policy $\pi_{\theta}$, old policy $\pi_{\theta_{\mathrm{old}}}$, frozen teacher $\pi_{\phi}$, group size $G$, sensitivity $\beta_h$, distillation coefficient $\lambda_{\mathrm{OPD}}$, log-ratio bound $c$
\ENSURE Optimized student policy $\pi_{\theta}$

\FOR{each training step}
    \STATE Sample $G$ trajectories $\{\tau_i,R_i,y_i\}_{i=1}^{G} \sim \pi_{\theta_{\mathrm{old}}}$ for each task

    \STATE GRPO advantage: normalize rewards within each trajectory group,
    \STATE \quad $A_i \leftarrow \dfrac{R_i-\frac{1}{G}\sum_{j=1}^{G}R_j}{\operatorname{std}(\{R_j\}_{j=1}^{G})+\epsilon}$

    \STATE Hindsight skill induction: for each trajectory $\tau_i$,
    \STATE \quad $z_i \leftarrow G_{\phi}(\tau_i,y_i)$

    \STATE Outcome-asymmetric credit assignment: for each decision turn $(i,t)$,
    \STATE \quad Compute teacher probabilities $q_{\phi}^{0}(\cdot\mid s_{i,t})$ and $q_{\phi}^{z_i}(\cdot\mid s_{i,t})$
    \STATE \quad $\Delta_{i,t,k} \leftarrow \log q_{\phi}^{z_i}(a_{i,t,k}\mid s_{i,t})-\log q_{\phi}^{0}(a_{i,t,k}\mid s_{i,t})$
    \STATE \quad $c_{i,t,k} \leftarrow m_{i,t,k}\!\left[2\sigma\!\left(\beta_h(2y_i-1)\Delta_{i,t,k}\right)-1\right]_{+}$
    \STATE \quad $H_{i,t} \leftarrow \dfrac{\sum_k c_{i,t,k}}{\max(1,\sum_k m_{i,t,k})}$

    \STATE Action-level on-policy distillation:
    \STATE \quad $\ell_{i,t,k} \leftarrow \log q_{\phi}^{z_i}(a_{i,t,k}\mid s_{i,t})-\log\pi_{\theta}(a_{i,t,k}\mid s_{i,t})$
    \STATE \quad $r_{i,t,k} \leftarrow \exp\!\left(\operatorname{clip}(\ell_{i,t,k},-c,c)\right)$
    \STATE \quad $\widehat{\mathcal{L}}_{\mathrm{ToolOPD}} \leftarrow
    \dfrac{1}{N_A}\sum_{i,t,k}m_{i,t,k}H_{i,t}
    \left(r_{i,t,k}\log r_{i,t,k}-r_{i,t,k}+1\right)$

    \STATE GRPO policy loss:
    \STATE \quad $\mathcal{L}_{\mathrm{GRPO}} \leftarrow
    -\mathbb{E}_{i,t,k}\!\left[
    \min\!\left(
    \rho_{i,t,k}A_i,\,
    \operatorname{clip}(\rho_{i,t,k},1-\epsilon_l,1+\epsilon_h)A_i
    \right)\right]$
    \STATE \quad where $\rho_{i,t,k}=
    \pi_{\theta}(a_{i,t,k}\mid s_{i,t})/
    \pi_{\theta_{\mathrm{old}}}(a_{i,t,k}\mid s_{i,t})$

    \STATE Joint policy update:
    \STATE \quad $\mathcal{L}_{\mathrm{total}} \leftarrow
    \mathcal{L}_{\mathrm{GRPO}}
    +\lambda_{\mathrm{OPD}}\widehat{\mathcal{L}}_{\mathrm{ToolOPD}}$
    \STATE \quad $\theta \leftarrow
    \theta-\eta\nabla_{\theta}\mathcal{L}_{\mathrm{total}}$
\ENDFOR

\RETURN $\pi_{\theta}$
\end{algorithmic}
\end{algorithm}

\definecolor{toolopdpurple}{HTML}{8064A2}
\definecolor{toolopdlightpurple}{HTML}{F4EFF8}
\newtcolorbox{trajectorybox}[1]{
    enhanced,
    breakable,
    colback=white,
    colframe=toolopdpurple,
    colbacktitle=toolopdpurple,
    coltitle=white,
    fonttitle=\bfseries,
    title={#1},
    boxrule=0.7pt,
    arc=2mm,
    outer arc=2mm,
    left=2.5mm,
    right=2.5mm,
    top=1.8mm,
    bottom=1.8mm,
    before skip=6pt,
    after skip=6pt
}

\newtcolorbox{skillbox}{
    enhanced,
    breakable,
    colback=toolopdlightpurple,
    colframe=toolopdpurple,
    colbacktitle=toolopdpurple,
    coltitle=white,
    fonttitle=\bfseries,
    title={Generated Hindsight Skill},
    boxrule=0.7pt,
    arc=2mm,
    outer arc=2mm,
    left=2.5mm,
    right=2.5mm,
    top=1.8mm,
    bottom=1.8mm,
    before skip=6pt,
    after skip=6pt
}

\definecolor{toolopdpurple}{HTML}{8064A2}

\newtcolorbox{casebox}[1]{
    enhanced,
    breakable,
    colback=white,
    colframe=toolopdpurple,
    colbacktitle=toolopdpurple,
    coltitle=white,
    fonttitle=\bfseries,
    title={#1},
    boxrule=0.7pt,
    arc=2mm,
    outer arc=2mm,
    left=3mm,
    right=3mm,
    top=2.5mm,
    bottom=2.5mm,
    before skip=8pt,
    after skip=8pt
}

\section{Trajectory Examples and Skill Generation}
\label{ap:trajectory_examples}

We present representative trajectories from ALFWorld and FTRL to illustrate how
ToolOPD induces reusable hindsight skills from completed interactions. We retain
all outcome-critical tool actions while omitting repetitive reasoning and
redundant observations using $[\ldots]$.

\begin{casebox}{Case 1: Long-Horizon Interaction in ALFWorld}

\textbf{Task.}
Put a clean cup in the microwave.

\medskip
\textbf{Trajectory.}

\textbf{Agent -- Step 1:}
\texttt{go to microwave 1}

\textbf{Environment -- Step 1:}
The agent arrives at \texttt{microwave 1}, which is closed.

\smallskip
\textbf{Agent -- Step 2:}
\texttt{open microwave 1}

\textbf{Environment -- Step 2:}
The microwave is open and contains \texttt{cup 1}.

\smallskip
\textbf{Agent and Environment -- Steps 3--11:}
The agent repeatedly examines the microwave and checks its inventory, while the
cup remains inside the microwave. $[\ldots]$

\smallskip
\textbf{Agent -- Step 12:}
\textbf{\texttt{take cup 1 from microwave 1}}

\textbf{Environment -- Step 12:}
The agent successfully picks up \texttt{cup 1}.

\smallskip
\textbf{Agent -- Step 13:}
\texttt{examine cup 1}

\textbf{Environment -- Step 13:}
The cup is empty, but its required clean state has not yet been established.

\smallskip
\textbf{Agent and Environment -- Steps 14--18:}
The agent performs several redundant examinations and location transitions that
do not change the task state. $[\ldots]$

\smallskip
\textbf{Agent -- Steps 19--20:}

\texttt{go to sinkbasin 1}

\textbf{\texttt{clean cup 1 with sinkbasin 1}}

\textbf{Environment -- Step 20:}
The cup is successfully cleaned.

\smallskip
\textbf{Agent -- Steps 21--24:}

\texttt{go to microwave 1}

\texttt{examine cup 1}

\texttt{examine microwave 1}

\textbf{\texttt{move cup 1 to microwave 1}}

\textbf{Environment -- Step 24:}
The clean cup is placed in the microwave, completing the task.

\medskip
\textbf{Outcome.}
Success.

\medskip
\textbf{Generated Hindsight Skill.}
\second{Verify and clean objects before placing them in appliances.}

\end{casebox}

Although the trajectory contains numerous examinations and unnecessary
transitions, its success is primarily determined by the pivotal action sequence
\textbf{\texttt{take} $\rightarrow$ \texttt{clean} $\rightarrow$
\texttt{place}}. ToolOPD therefore extracts an outcome-grounded skill from the
noisy trajectory and concentrates supervision on actions that causally change
the task state.

\begin{casebox}{Case 2: Multi-Step Tool Reasoning on FTRL}

\textbf{Task.}
What is the result of subtracting 8 from the reverse of the birth year of the
star of \textit{Lady in a Cage}?

\medskip
\textbf{Trajectory.}

\textbf{Agent -- Step 1:}
The agent identifies the main star of the target film:

{\bfseries\ttfamily
movie\_star\_finder(\\
\hspace*{1em}movie\_title="Lady in a Cage",\\
\hspace*{1em}role\_type="main\_star")
}

\textbf{Environment -- Step 1:}
The tool returns \textbf{\texttt{Olivia de Havilland}}.

\smallskip
\textbf{Agent -- Step 2:}
The agent retrieves the birth year of the identified person:

{\bfseries\ttfamily
biographical\_info\_retriever(\\
\hspace*{1em}name="Olivia de Havilland",\\
\hspace*{1em}information\_type="birth\_year")
}

\textbf{Environment -- Step 2:}
The tool returns \textbf{\texttt{1916}}.

\smallskip
\textbf{Agent -- Step 3:}
The agent reverses \textbf{\texttt{1916}} to obtain
\textbf{\texttt{6191}} and performs the required subtraction:

{\bfseries\ttfamily
advanced\_arithmetic\_calculator(\\
\hspace*{1em}operation="subtract",\\
\hspace*{1em}operands=[6191, 8])
}

\textbf{Environment -- Step 3:}
The tool returns \textbf{\texttt{6183}}.

\smallskip
\textbf{Agent -- Final Answer:}
\textbf{\texttt{<answer>6183</answer>}}

\medskip
\textbf{Outcome.}
Correct answer.

\medskip
\textbf{Generated Hindsight Skill.}
Resolve entities before computation: retrieve each dependent entity and
numerical attribute with tools, and then apply the requested transformations in
their specified order.

\end{casebox}

The FTRL trajectory forms a clear dependency chain: identifying the target
person and retrieving the corresponding birth year are necessary prerequisites
for the final numerical operation. Conditioned on the successful outcome,
ToolOPD preserves supervision along this reasoning chain and emphasizes the tool
calls that establish essential intermediate variables. Together, the two cases
show that ToolOPD induces transferable skills and assigns credit to
outcome-relevant decisions across embodied interaction and multi-step tool
reasoning.


\definecolor{ToolOPDPurple}{HTML}{8064A2}
\definecolor{PromptBackground}{HTML}{F7F4FA}

\newtcblisting{promptbox}[1]{
    enhanced,
    breakable,
    colback=PromptBackground,
    colframe=ToolOPDPurple,
    colbacktitle=ToolOPDPurple,
    coltitle=white,
    fonttitle=\bfseries,
    title={#1},
    boxrule=0.7pt,
    arc=2mm,
    left=1.2mm,
    right=1.2mm,
    top=1mm,
    bottom=1mm,
    listing only,
    listing options={
        basicstyle=\ttfamily\footnotesize,
        breaklines=true,
        columns=fullflexible,
        keepspaces=true,
        showstringspaces=false
    }
}

\section{Prompts Used for Training}
\label{ap:training_prompts}

We report the prompts used to collect on-policy trajectories and construct
ToolOPD supervision. For environment interaction, each prompt specifies the
task, recent interaction history, current observation, admissible actions or
tools, and the required response format. For ToolOPD, the completed trajectory
and its outcome are additionally provided to the teacher for hindsight skill
induction. The generated skill is subsequently inserted into the teacher
context when computing action-level distillation targets. Placeholders enclosed
in braces are instantiated online.

\subsection{Environment Interaction Prompts}

\subsubsection{ALFWorld}

\begin{promptbox}{ALFWorld Interaction Prompt}
You are an expert agent operating in the ALFRED Embodied Environment.

Your task is to:
{TASK_DESCRIPTION}

Prior to this step, you have already taken {STEP_COUNT} step(s).
Below are the most recent {HISTORY_LENGTH} observations and
the corresponding actions you took:
{ACTION_HISTORY}

You are now at step {CURRENT_STEP}.
Your current observation is:
{CURRENT_OBSERVATION}

Your admissible actions in the current situation are:
[{ADMISSIBLE_ACTIONS}]

Now it is your turn to take one action. First reason step by step
about the current situation. The reasoning must be enclosed within
<think> and </think> tags and contain at most two sentences.

Then select exactly one admissible action and present it within
<action> and </action> tags. Output one action only and then stop.
\end{promptbox}

For the first interaction step, the history block is omitted because no
previous observation--action pair is available.

\subsubsection{WebShop}

\begin{promptbox}{WebShop Interaction Prompt}
You are an expert autonomous agent operating in the WebShop
e-commerce environment.

Your task is to:
{TASK_DESCRIPTION}

Prior to this step, you have already taken {STEP_COUNT} step(s).
Below are the most recent {HISTORY_LENGTH} observations and
the corresponding actions you took:
{ACTION_HISTORY}

You are now at step {CURRENT_STEP}.
Your current observation is:
{CURRENT_OBSERVATION}

Your admissible actions in the current situation are:
[
{AVAILABLE_ACTIONS}
]

Now it is your turn to take one action. First reason step by step
about the current situation and determine which admissible action
best advances the shopping goal. The reasoning must be enclosed
within <think> and </think> tags.

Then select one admissible action and present it within
<action> and </action> tags.
\end{promptbox}

As in ALFWorld, the interaction-history block is omitted at the first step.

\subsubsection{FTRL}

For FTRL, the available tool schemas vary across instances and are dynamically
inserted into the system prompt.

\begin{promptbox}{FTRL System Prompt}
# Tools

You may call one or more functions to assist with the user query.

You are provided with function signatures within <tools></tools>
XML tags:

<tools>
{TOOL_SCHEMAS}
</tools>

For each function call, return a JSON object containing the
function name and arguments within <tool_call></tool_call> tags:

<tool_call>
{"name": <FUNCTION_NAME>, "arguments": <ARGUMENTS_JSON>}
</tool_call>
\end{promptbox}

\begin{promptbox}{FTRL User Prompt}
Please call the given tools to answer the question. All information
must be obtained by calling tools rather than answering the question
directly.

If a tool call fails, correct the call and continue until you arrive
at an answer. Output only the final answer, as a word, number, or
short phrase, inside the <answer></answer> tag, without explanations
or additional information.

Question:
{QUESTION}
\end{promptbox}

\subsection{Outcome-Conditioned Hindsight Skill Induction}

ToolOPD applies an independent teacher to the complete trajectory after its
outcome becomes available. The prompt explicitly distinguishes successful and
failed trajectories so that the generated skill captures either an effective
workflow or an error-avoidance principle.

\subsubsection{ALFWorld and WebShop}

\begin{promptbox}{Hindsight Skill Induction Prompt}
You are an independent RL teacher analyzing a completed
long-horizon tool-use trajectory.

Task description:
{TASK_DESCRIPTION}

Complete trajectory:
{COMPLETE_TRAJECTORY}

Final outcome:
{SUCCESS_OR_FAILURE}

If the trajectory succeeds:
Describe the reusable workflow that made the trajectory succeed.

If the trajectory fails:
Identify the task-specific failed decision pattern and state one
reusable workflow or avoidance principle. Ground every detail in
the task and trajectory. Explicitly identify the target object,
the relevant receptacle or tool, the missing prerequisite operation
or subgoal, and any wrong object selection or repeated no-progress
action loop. For tasks requiring cleaning, cooling, or heating,
distinguish actively performing the operation from merely finding
an object that already has the desired state. Do not invent
unobserved objects or states.

Return ONLY one valid JSON object:
{"skill": "one concise policy-facing skill"}

Do not output turn numbers, importance scores, action corrections,
numeric ratings, or Yes/No judgments.
\end{promptbox}

\subsubsection{FTRL}

\begin{promptbox}{FTRL Hindsight Skill Induction Prompt}
You are an expert tool-use teacher.

Trajectory:
{COMPLETE_TRAJECTORY}

Outcome:
{SUCCESS_OR_FAILURE}

Induce exactly one short, reusable tool-use strategy.

For a successful trajectory, capture the effective workflow or
pivotal decision. For a failed trajectory, capture the root error
and explain how to avoid it.

Use exactly the following format:
<skill>
Rule Name: imperative guidance. Apply when the relevant condition holds.
</skill>

Use at most 50 words. Do not include analysis, scores, or a
replacement trajectory.
\end{promptbox}

\subsection{Skill-Conditioned Teacher Prompt}

After hindsight skill induction, ToolOPD prepends the generated skill to the
original user context of the teacher. The remaining environment prompt, tool
schemas, and interaction history are kept unchanged.

\begin{promptbox}{Skill-Conditioned Teacher Context}
[Hindsight Skill]
{GENERATED_SKILL}

{ORIGINAL_USER_PROMPT}
\end{promptbox}

The student policy receives the original environment prompt, whereas the
teacher is conditioned on the generated hindsight skill. ToolOPD compares their
action-token distributions and uses the resulting outcome-asymmetric skill
relevance to dynamically weight the distillation objective.

\section{Empirical Study: Detailed Setup}
\label{ap:empirical_study_settings}
To investigate the connection between performance degradation in long-horizon tool learning and outcome-dependent credit assignment, we design three complementary diagnostic experiments. These experiments examine long-context robustness, outcome sensitivity, and supervision conflicts (Figure~1(a)--(c)). The first evaluates the impact of interaction length on model performance, while the latter two analyze the behavior of ToolOPD credit weights on fixed trajectories. We uniformly use Qwen3-1.7B for the long-context experiments. For the credit analysis, we use Qwen3-4B models trained up to step 150 on each dataset as the teacher and probability evaluator.

\textbf{Credit concentration.}
Fig.~\ref{fig:motivation}(a) examines how credit is distributed across interaction turns in long-horizon ALFWorld trajectories. For each trajectory, we rank all turns according to their estimated outcome-aware credit $H_{i,t}$ and compute the fraction of total credit assigned to the top $20\%$ of turns:
\[
\mathrm{CreditMass}_{20\%}
=
\frac{
\sum_{t \in \mathcal{T}^{i}_{20\%}} H_{i,t}
}{
\sum_{t \in \mathcal{T}^{i}} H_{i,t}
},
\]
where $\mathcal{T}^{i}_{20\%}$ denotes the highest-credit $20\%$ of turns in trajectory $i$. As shown in Fig.~\ref{fig:motivation}(a), these turns account for $74.2\%$ of the total credit, whereas the remaining $80\%$ account for only $25.8\%$. This highly concentrated distribution shows that different interaction turns contribute unequally to the final outcome. Consequently, uniform OPD spreads teacher supervision over many low-credit transitions and weakens the learning signal for pivotal tool decisions.

\textbf{Outcome-sensitive credit assignment.} Fig.~\ref{fig:motivation} (b) employs a controlled outcome counterfactual experiment. We keep the task, state, and action sequence unchanged. We then flip the final outcome between success and failure, and regenerate the factual and counterfactual skills accordingly. Next, we compute the factual credit $H_{i,t}$ and the counterfactual credit $H^{\mathrm{cf}}_{i,t}$ on the same action tokens. The credit calculation follows the outcome-asymmetric definition of ToolOPD:
\[
H_{i,t,k}
=
\left[
2\sigma\!\left(
\beta s_i
\bigl(
\log \pi_{\mathrm{tea}}(a_{i,t,k}\mid z_i)
-
\log \pi_{\mathrm{tea}}(a_{i,t,k})
\bigr)
\right)-1
\right]_{+},
\]
where $s_i \in \{1, -1\}$ denotes success and failure, respectively, and we set $\beta=5$. If a turn satisfies $\left|H_{i,t}-H^{\mathrm{cf}}_{i,t}\right|>0.01$, we label it as an outcome-sensitive turn. This experiment covers sampled data from ALFWorld (6 trajectories, 29 turns), WebShop (4 trajectories, 12 turns), and SearchQA (4 trajectories, 8 turns). It directly quantifies the dependence of credit assignment on the final outcome.

\textbf{Conflicting supervision analysis.} Fig.~\ref{fig:motivation} (c) analyzes the consistency between the evidence provided by the teacher skill for observed actions and the final outcome. For an action token $a_{i,t,k}$, we define the signed teacher evidence as:
\[
e_{i,t,k}
=
s_i
\left[
\log \pi_{\mathrm{tea}}(a_{i,t,k}\mid z_i)
-
\log \pi_{\mathrm{tea}}(a_{i,t,k})
\right].
\]
When $e_{i,t,k}>0$, skill conditioning shifts the teacher distribution in a direction consistent with the outcome; otherwise, it conflicts. We compute the proportion of conflicting evidence in both uniform distillation and credit-weighted distillation:
\[
R_{\mathrm{uniform}}
=
\frac{\sum_{i,t,k}[-e_{i,t,k}]_{+}}
     {\sum_{i,t,k}|e_{i,t,k}|},
\qquad
R_{H}
=
\frac{\sum_{i,t,k}H_{i,t}[-e_{i,t,k}]_{+}}
     {\sum_{i,t,k}H_{i,t}|e_{i,t,k}|}.
\]
Both settings share the same trajectories and teacher outputs. The only difference is whether ToolOPD credit weighting is applied. We conduct this analysis on ALFWorld diagnostic trajectories and report the conflict proportions for successful, failed, and all trajectories.

Additionally, Fig.~\ref{fig:motivation} (d) shows the success rate of Qwen3-0.6B on ALFWorld over 150 continuous training steps. This verifies whether the differences in the aforementioned credit assignment mechanisms translate into actual optimization gains.

\begin{table*}[t]
    \centering
    \caption{Task and metric abbreviations used in Table~\ref{tab:horizon_results}.}
    \label{tab:task_abbreviations}
    \small
    \setlength{\tabcolsep}{7pt}
    \renewcommand{\arraystretch}{1.08}
    \begin{tabular}{ll ll ll}
        \toprule
        \multicolumn{2}{c}{\textbf{ALFWorld}}
        & \multicolumn{2}{c}{\textbf{SearchQA}}
        & \multicolumn{2}{c}{\textbf{WebShop}} \\
        \cmidrule(lr){1-2}
        \cmidrule(lr){3-4}
        \cmidrule(lr){5-6}
        \textbf{Abbr.} & \textbf{Full name}
        & \textbf{Abbr.} & \textbf{Full name}
        & \textbf{Abbr.} & \textbf{Full name} \\
        \midrule
        Pick  & Pick and Place
        & NQ   & Natural Questions
        & Score & Average Score \\

        Look  & Look at Object in Light
        & Triv & TriviaQA
        & Succ. & Success Rate \\

        Clean & Pick, Clean, and Place
        & Pop  & PopQA
        & & \\

        Heat  & Pick, Heat, and Place
        & Hotp & HotpotQA
        & & \\

        Cool  & Pick, Cool, and Place
        & 2Wk  & 2WikiMultiHopQA
        & & \\

        Pick2 & Pick Two Objects and Place
        & MuS  & MuSiQue
        & & \\

        & & Bam & Bamboogle
        & & \\
        \bottomrule
    \end{tabular}

    \vspace{2pt}
    \footnotesize
\end{table*}

\begin{table*}[t]
    \centering
    \caption{Metric abbreviations used in the FTRL, BFCL, and ToolHop evaluation.}
    \label{tab:ftrl_metric_abbreviations}
    \small
    \setlength{\tabcolsep}{7pt}
    \renewcommand{\arraystretch}{1.08}
    \begin{tabular}{ll ll ll}
        \toprule
        \multicolumn{2}{c}{\textbf{FTRL}}
        & \multicolumn{2}{c}{\textbf{BFCL}}
        & \multicolumn{2}{c}{\textbf{ToolHop}} \\
        \cmidrule(lr){1-2}
        \cmidrule(lr){3-4}
        \cmidrule(lr){5-6}
        \textbf{Abbr.} & \textbf{Full name}
        & \textbf{Abbr.} & \textbf{Full name}
        & \textbf{Abbr.} & \textbf{Full name} \\
        \midrule
        Solve-F1 & Solve F1 Score
        & Base & Multi-Turn Base
        & Acc. & Accuracy \\

        Solve-R & Solve Recall
        & MF & Multi-Turn Missing Function
        & & \\

        Solve-P & Solve Precision
        & MP & Multi-Turn Missing Parameter
        & & \\

        & & LC & Multi-Turn Long Context
        & & \\

        & & MTOA & Multi-Turn Overall Accuracy
        & & \\

        & & Memory & Agent Memory Accuracy
        & & \\
        \bottomrule
    \end{tabular}

    \vspace{2pt}
\end{table*}

\section{Evaluation metrics}
\label{ap:evl_metric}
\section{Evaluation metrics}
\label{ap:eval_metrics}
For ALFWorld, SearchQA, and WebShop in Table~\ref{tab:horizon_results}, we use task success rate as the primary metric, defined as the proportion of evaluation instances in which the agent successfully completes the environment objective or produces the correct final answer. WebShop additionally reports a normalized task score that measures partial progress in unsuccessful trajectories. The task abbreviations are summarized in Table~\ref{tab:task_abbreviations}. The column ``Avg.'' denotes the average performance across the corresponding task categories.

For Table~\ref{tab:tool-use_agent_results}, we follow the official evaluation protocol of each benchmark. FTRL reports Solve-P, Solve-R, and Solve-F1, which measure tool-use precision, task completion, and their harmonic mean, respectively. BFCL reports accuracy on multi-turn base, missing-function, missing-parameter, long-context, and memory tasks, with MTOA denoting Multi-Turn Overall Accuracy. ToolHop uses Accuracy (Acc.) to evaluate the correctness of the final answer after multi-hop tool reasoning. Unless otherwise specified, all metrics are reported as percentages, and higher values indicate better performance. The corresponding abbreviations are summarized in Table~\ref{tab:ftrl_metric_abbreviations}.
\end{document}